\documentclass{fairmeta}
\setcitestyle{numbers,square}
\usepackage{amsmath,amssymb,amsfonts}
\usepackage{graphicx,multirow,booktabs,tabularx,array}
\usepackage{xcolor,textcomp,pdflscape}

\usepackage{pifont}
\usepackage{fontawesome5}
\newcolumntype{x}[1]{>{\centering\arraybackslash}p{#1}}
\newcolumntype{y}[1]{>{\raggedright\arraybackslash}p{#1}}
\newcolumntype{z}[1]{>{\raggedleft\arraybackslash}p{#1}}
\newcolumntype{M}[1]{>{\raggedright\arraybackslash}m{#1}}
\newcolumntype{C}[1]{>{\centering\arraybackslash}m{#1}}
\newcommand{\tablestyle}[2]{\setlength{\tabcolsep}{#1}\renewcommand{\arraystretch}{#2}\centering\footnotesize}

\newcommand{\SurveyView}[1]{\textsf{#1}}
\DeclareRobustCommand{\ResourceModality}[1]{\mbox{\fontsize{7.5}{8.5}\selectfont\makebox[1.3em][c]{\textcolor{black!75}{#1}}}}
\DeclareRobustCommand{\ModDepth}{\ResourceModality{\faLayerGroup}}
\DeclareRobustCommand{\ModMono}{\ResourceModality{\faAdjust}}
\DeclareRobustCommand{\ModAudio}{\ResourceModality{\faVolumeUp}}
\DeclareRobustCommand{\ModIMU}{\ResourceModality{\faCompass}}
\DeclareRobustCommand{\ModGaze}{\ResourceModality{\faEye}}
\newcolumntype{Y}{>{\raggedright\arraybackslash}X}

\definecolor{tableyes}{HTML}{2E7D32}
\definecolor{tableno}{HTML}{A02337}
\newcommand{\tablesymbol}[2]{\mbox{\fontsize{8}{9}\selectfont\makebox[1em][c]{\textcolor{#1}{\ding{#2}}}}}
\DeclareRobustCommand{\Yes}{\tablesymbol{tableyes}{51}}
\DeclareRobustCommand{\No}{\tablesymbol{tableno}{55}}
\usepackage{tikz}

\newsavebox{\LAIconReferenceBox}
\DeclareRobustCommand{\SurveyVectorIcon}[2][\Yes{}]{%
  \mbox{\begingroup
    \sbox{\LAIconReferenceBox}{#1}%
    \raisebox{-\dp\LAIconReferenceBox}{%
      \resizebox*{!}{\dimexpr\ht\LAIconReferenceBox+\dp\LAIconReferenceBox\relax}{%
        \tikz[x=6pt,y=6pt,baseline=(current bounding box.south),
              draw=black!75,fill=black!75,line width=0.42pt,
              line cap=round,line join=round]{#2}%
      }%
    }%
  \endgroup}%
}
\DeclareRobustCommand{\LAProvenanceIcon}[1]{\SurveyVectorIcon{#1}}

\DeclareRobustCommand{\LAHumanIcon}{%
  \LAProvenanceIcon{%
    \fill (0.50,0.78) circle[radius=0.21];
    \fill
      (0.05,0.03) -- (0.95,0.03) -- (0.95,0.15)
      .. controls (0.95,0.35) and (0.76,0.44) .. (0.63,0.47)
      -- (0.63,0.53) -- (0.37,0.53) -- (0.37,0.47)
      .. controls (0.24,0.44) and (0.05,0.35) .. (0.05,0.15)
      -- cycle;
  }%
}

\DeclareRobustCommand{\LAModelIcon}{%
  \LAProvenanceIcon{%
    \fill (0.46,0.76) rectangle (0.54,0.91);
    \fill (0.50,0.97) circle[radius=0.06];
    \fill[rounded corners=0.7pt] (0.13,0.10) rectangle (0.87,0.76);
    \fill[rounded corners=0.2pt] (0.03,0.34) rectangle (0.15,0.57);
    \fill[rounded corners=0.2pt] (0.85,0.34) rectangle (0.97,0.57);
    \fill[white] (0.34,0.51) circle[radius=0.085];
    \fill[white] (0.66,0.51) circle[radius=0.085];
    \fill[white,rounded corners=0.15pt] (0.30,0.25) rectangle (0.70,0.33);
  }%
}

\DeclareRobustCommand{\LAUnknownIcon}{%
  \LAProvenanceIcon{%
    \fill (0.50,0.50) circle[radius=0.46];
    \fill[white] (0.29,0.66)
      .. controls (0.29,0.80) and (0.39,0.86) .. (0.52,0.86)
      .. controls (0.67,0.86) and (0.77,0.77) .. (0.77,0.65)
      .. controls (0.77,0.54) and (0.69,0.49) .. (0.62,0.44)
      .. controls (0.58,0.41) and (0.57,0.39) .. (0.57,0.35)
      -- (0.43,0.35)
      .. controls (0.43,0.44) and (0.48,0.48) .. (0.54,0.52)
      .. controls (0.59,0.56) and (0.63,0.59) .. (0.63,0.65)
      .. controls (0.63,0.70) and (0.59,0.73) .. (0.52,0.73)
      .. controls (0.46,0.73) and (0.43,0.70) .. (0.43,0.66)
      -- cycle;
    \fill[white] (0.50,0.22) circle[radius=0.07];
  }%
}

\DeclareRobustCommand{\LAStatus}[1]{\mbox{\makebox[9.4mm][l]{#1}}}
\DeclareRobustCommand{\LANone}{\LAStatus{\No{}}}
\DeclareRobustCommand{\LAHuman}{\LAStatus{\Yes{}\kern0.1em\LAHumanIcon{}}}
\DeclareRobustCommand{\LAModel}{\LAStatus{\Yes{}\kern0.1em\LAModelIcon{}}}
\DeclareRobustCommand{\LAMixed}{%
  \LAStatus{\Yes{}\kern0.1em\LAHumanIcon{}\kern0.08em\LAModelIcon{}}}
\DeclareRobustCommand{\LAUnknown}{%
  \LAStatus{\Yes{}\kern0.1em\LAUnknownIcon{}}}
\DeclareRobustCommand{\LAAutoUnknown}{%
  \LAStatus{\Yes{}\kern0.1em\LAUnknownIcon{}\textsuperscript{a}}}

\DeclareRobustCommand{\AnnotationIcon}[1]{\SurveyVectorIcon[\ModDepth{}]{#1}}

\DeclareRobustCommand{\AnnHand}{%
  \AnnotationIcon{%
    \fill (0.20,0.00) -- (0.81,0.00) -- (0.81,0.14)
      .. controls (1.00,0.30) and (1.04,0.42) .. (1.04,0.52)
      -- (1.04,0.70) arc[start angle=0,end angle=180,radius=0.08]
      -- (0.88,0.49) -- (0.80,0.49) -- (0.80,0.90)
      arc[start angle=0,end angle=180,radius=0.08]
      -- (0.64,0.49) -- (0.56,0.49) -- (0.56,1.02)
      arc[start angle=0,end angle=180,radius=0.08]
      -- (0.40,0.49) -- (0.32,0.49) -- (0.32,0.93)
      arc[start angle=0,end angle=180,radius=0.08]
      -- (0.16,0.40) -- (0.03,0.54)
      .. controls (-0.12,0.70) and (-0.29,0.56) .. (-0.17,0.40)
      -- (0.11,0.16)
      .. controls (0.20,0.10) and (0.20,0.05) .. (0.20,0.00)
      -- cycle;
  }%
}

\DeclareRobustCommand{\AnnObject}{%
  \AnnotationIcon{%
    \fill (0.00,0.77) -- (0.50,1.02) -- (1.00,0.77) -- (0.50,0.52) -- cycle;
    \fill (0.00,0.70) -- (0.46,0.47) -- (0.46,0.00) -- (0.00,0.23) -- cycle;
    \fill (0.54,0.47) -- (1.00,0.70) -- (1.00,0.23) -- (0.54,0.00) -- cycle;
  }%
}

\DeclareRobustCommand{\AnnTask}{%
  \AnnotationIcon{%
    \draw[line width=0.65pt] (0.57,0.80) -- (0.57,0.54);
    \draw[line width=0.65pt] (0.18,0.24) -- (0.18,0.54)
      -- (0.96,0.54) -- (0.96,0.24);
    \fill[rounded corners=0.25pt] (0.39,0.76) rectangle (0.75,1.03);
    \fill[rounded corners=0.25pt] (0.00,0.00) rectangle (0.36,0.27);
    \fill[rounded corners=0.25pt] (0.78,0.00) rectangle (1.14,0.27);
  }%
}

\DeclareRobustCommand{\AnnContact}{%
  \AnnotationIcon{%
    \draw[line width=0.65pt] (0.40,1.40) circle[radius=0.26];
    \fill (0.31,0.00) -- (0.83,0.00) -- (0.84,0.20)
      .. controls (0.99,0.43) and (0.98,0.52) .. (0.94,0.61)
      -- (0.94,0.71) arc[start angle=0,end angle=180,radius=0.07]
      -- (0.80,0.64) -- (0.80,0.79)
      arc[start angle=0,end angle=180,radius=0.07]
      -- (0.66,0.67) -- (0.66,0.86)
      arc[start angle=0,end angle=180,radius=0.08]
      -- (0.50,1.27) arc[start angle=0,end angle=180,radius=0.10]
      -- (0.30,0.63) -- (0.17,0.79)
      .. controls (0.06,0.91) and (-0.08,0.77) .. (0.01,0.64)
      -- (0.22,0.28)
      .. controls (0.28,0.20) and (0.31,0.12) .. (0.31,0.00)
      -- cycle;
  }%
}

\DeclareRobustCommand{\AnnSemantic}{%
  \AnnotationIcon{%
    \fill[rounded corners=0.35pt] (0.03,0.53) -- (0.56,1.06)
      -- (1.08,1.06) -- (1.08,0.54) -- (0.55,0.01) -- cycle;
    \fill[white] (0.87,0.85) circle[radius=0.10];
  }%
}

\DeclareRobustCommand{\AnnotationSlot}[1]{%
  \mbox{\fontsize{7.5}{8.5}\selectfont\makebox[1.3em][c]{#1}}%
}
\DeclareRobustCommand{\AnnTypes}[5]{%
  \mbox{%
    \ifnum\numexpr#1+#2+#3+#4+#5\relax=0
      -%
    \else
      \ifnum#1=1\relax\AnnotationSlot{\AnnHand{}}\fi%
      \ifnum#2=1\relax\AnnotationSlot{\AnnObject{}}\fi%
      \ifnum#3=1\relax\AnnotationSlot{\AnnTask{}}\fi%
      \ifnum#4=1\relax\AnnotationSlot{\AnnContact{}}\fi%
      \ifnum#5=1\relax\AnnotationSlot{\AnnSemantic{}}\fi%
    \fi
  }%
}

\DeclareUnicodeCharacter{2713}{\Yes}
\DeclareUnicodeCharacter{25CB}{\ensuremath{\circ}}
\DeclareUnicodeCharacter{2717}{\No}
\DeclareUnicodeCharacter{03C4}{\ensuremath{\tau}}
\DeclareUnicodeCharacter{03C0}{\ensuremath{\pi}}

\newsavebox{\milestonediagrambox}


\title{From Egocentric Human Video to Robot Manipulation in the Era of Embodied Intelligence}

\author[1,\dagger,*]{Yi Wang}
\author[1,\dagger]{Jingqiu Wu}
\author[1,\dagger]{Lei Yao}
\author[2]{Zhaoshuo Li}
\author[1]{Haoyu Zhu}
\author[1]{Yirui Zhou}
\author[1]{Huaiyuan Xu}
\author[1]{Ying~Zheng}
\author[1]{Chuang Yang}
\author[1]{Lap-Pui Chau}

\affiliation[1]{Department of EEE, The Hong Kong Polytechnic University}
\affiliation[2]{Cosmos Lab, NVIDIA}

\contribution[*]{Corresponding author}
\contribution[\dagger]{Equal contribution}

\correspondence{Yi Wang, \email{wang1241@e.ntu.edu.sg (yi-eie.wang@polyu.edu.hk)}}

\abstract{Egocentric human video (EHV) records close-range hand-object interaction (HOI) and the temporal structure of everyday activities, offering a potentially scalable source of experience for robot manipulation. Using this experience for robot control requires recovering useful supervision without recorded robot actions and accounting for differences in morphology, viewpoint, and action space. We define \textit{Ego2Robot} learning as the process of learning or adapting robot manipulation policies from EHV, organized into four learning paradigms: explicit embodiment transfer, explicit action and task supervision, egocentric video representation pretraining, and latent action and dynamics modeling. The four paradigms differ in what they derive from human video: robot-format demonstrations, interpretable supervision, visual representations, or predictive models of actions and dynamics. For each paradigm, we trace what EHV information is extracted, how robot policies use it, and when the additional data is required. Besides, we review the benchmark datasets that cover EHV, robot demonstrations, and synthetic and simulated resources. This paper links EHV data's contribution to the task information preserved and the robot-specific grounding used during training or deployment. Finally, we identify challenges in physical feasibility, agent coordination, in-context adaptation, controlled evaluation, and predictive control.
}
\metadata[Project]{\url{https://mikewu597.github.io/Ego2Robot_Learning/}}

\begin{document}
\maketitle

\section{Introduction}\label{sec_intro}

Embodied intelligence studies how agents perceive, learn, and act in the physical world, with robot manipulation focusing on interactions that change object states to achieve task goals. Robot manipulation policies demand robot-object interaction experience that covers different tasks, objects, and environments~\cite{sunderhauf_limits_2018}. Collecting this experience directly on robots requires physical platforms, repeated task execution, teleoperation, and scene setup. However, these collection requirements restrict both the scalability and variety of robot demonstrations available for policy learning~\cite{mandlekar_mimicgen_2023, kareer_egomimic_2024}. This limitation motivates supplementing robot demonstrations with hand-object interaction (HOI) videos.

Egocentric human video (EHV), recorded by human-worn cameras during everyday activities, provides such HOI experience and a potential of scalability and variety. Large-scale EHV datasets record human activities across tasks and environments without requiring a robot to perform each interaction~\cite{damen_rescaling_2020, grauman_ego4d_2022}. In an egocentric view, the camera often records the demonstrator's hands, manipulated objects, and their changing spatial relationships at close range~\cite{su_interformer_2026}. Its temporal structure reveals how interactions unfold, including action ordering, object state changes, and task progression. These spatial and temporal cues offer interaction priors and task information for learning visual representations and manipulation policies.

Human video leaves the robot's control problem underspecified. Specifically, human hands and robot end-effectors differ in morphology and motion capabilities. Human-worn and robot-mounted cameras also observe interactions from different viewpoints. Moreover, EHV does not record robot action commands, and visual observations alone cannot fully determine the forces and contact dynamics required for execution. Existing approaches attempted to exploit EHV through cross-domain alignment of human and robot demonstrations~\cite{kareer_egomimic_2024} and visual representation pretraining~\cite{nair_r3m_2022}. Recently, latent action pretraining derives action-like targets from visual transitions without recorded action labels~\cite{ye_latent_2025}. We use the term \emph{Ego2Robot learning} for learning manipulation-relevant knowledge from EHV and grounding it in the capabilities of the target robot.

\begin{figure}[t]
\centering
\begingroup
\sbox{\milestonediagrambox}{\includegraphics[width=1.08\textwidth]{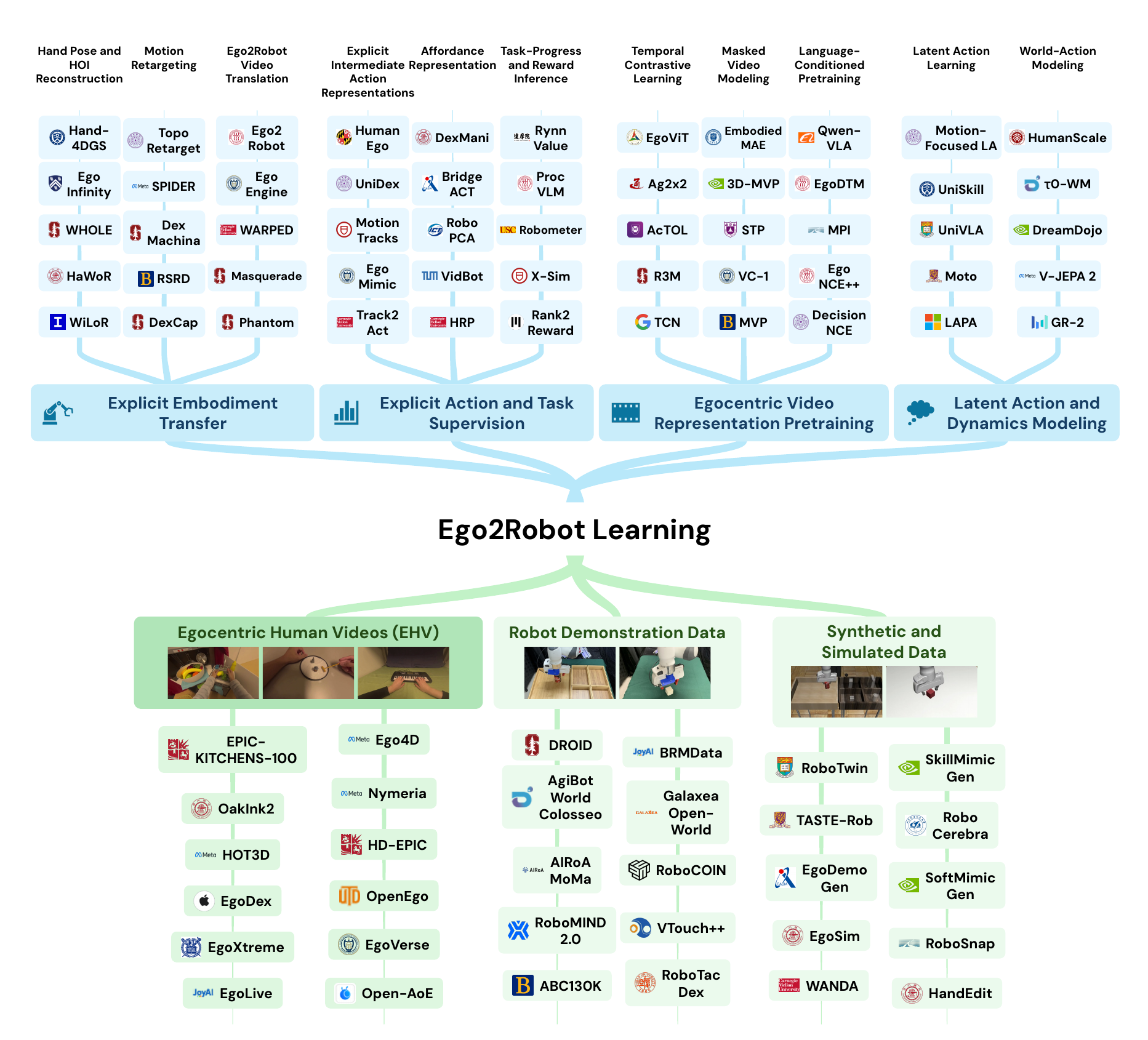}}%
\makebox[\textwidth][c]{%
\begingroup
\setlength{\unitlength}{\dimexpr\wd\milestonediagrambox/895\relax}%
\hypersetup{hidelinks}%
\ifdefined\pdflinkmargin
  \edef\milestonemargin{\the\pdflinkmargin}%
\else
  \def\milestonemargin{0pt}%
\fi
\def\milestonearea#1#2{%
  \phantom{\rule{\dimexpr#1\unitlength-\milestonemargin-\milestonemargin\relax}%
    {\dimexpr#2\unitlength-\milestonemargin-\milestonemargin\relax}}%
}%
\def\milestonehit#1#2#3#4#5{%
  \put(#1,#2){\kern\milestonemargin\raisebox{\milestonemargin}{%
    \hyperref[#5]{\milestonearea{#3}{#4}}}}%
}%
\def\milestoneanchorhit#1#2#3#4#5{%
  \put(#1,#2){\kern\milestonemargin\raisebox{\milestonemargin}{%
    \hyperlink{#5}{\milestonearea{#3}{#4}}}}%
}%
\rlap{\begin{picture}(0,0)%
  \milestonehit{24.000}{466.568}{221.000}{44.500}{sec_main_1_1}%
  \milestonehit{249.000}{466.568}{214.000}{44.500}{sec_main_1_2}%
  \milestonehit{467.000}{466.568}{226.000}{44.500}{sec_main_1_3}%
  \milestonehit{697.000}{466.568}{176.000}{44.500}{sec_main_1_4}%
  \milestonehit{104.500}{257.900}{272.000}{72.000}{sec_main_2_1}%
  \milestonehit{384.903}{257.900}{193.195}{72.000}{sec_main_2_2}%
  \milestoneanchorhit{602.500}{243.400}{174.000}{86.500}{milestone.synthetic-data}%
  \milestonehit{27.038}{742.410}{69.923}{35.318}{sec_taxonomy_hand_pose}%
  \milestonehit{101.337}{753.210}{53.327}{24.518}{sec_taxonomy_motion_retargeting}%
  \milestonehit{174.863}{742.410}{51.275}{35.318}{sec_taxonomy_video_translation}%
  \milestonehit{249.590}{731.610}{74.819}{46.118}{sec_taxonomy_intermediate_actions}%
  \milestonehit{327.241}{753.210}{51.518}{24.518}{sec_taxonomy_affordance}%
  \milestonehit{389.329}{742.410}{65.342}{35.318}{sec_taxonomy_task_reward}%
  \milestonehit{486.076}{742.410}{53.849}{35.318}{sec_taxonomy_temporal_contrastive}%
  \milestonehit{558.719}{742.410}{42.563}{35.318}{sec_taxonomy_masked_video}%
  \milestonehit{618.617}{742.410}{56.765}{35.318}{sec_taxonomy_language_pretraining}%
  \milestonehit{711.003}{753.210}{61.994}{24.518}{sec_taxonomy_latent_action}%
  \milestonehit{793.976}{753.210}{62.048}{24.518}{sec_taxonomy_world_action}%
\end{picture}}%
\endgroup%
%
    \usebox{\milestonediagrambox}%
}
\endgroup
\caption{Taxonomy of Ego2Robot learning of four learning paradigms with representative methods and supporting data resources. Example images from EgoDex~\cite{hoque_egodex_2026}, RoboMimic~\cite{robomimic}, and RoboTurk~\cite{roboturk}.}
\label{fig:taxonomy}
\end{figure}

\looseness=-1
Existing reviews address robot learning from robot demonstrations, general human videos, and embodied manipulation~\cite{correia_survey_2023, eze_learning_2025, ma_robot_2026, ye_data_2026, zheng_survey_2025}, where the comparisons with ours are shown in Table~\ref{tab:survey_comparison}. Recently, Ma et al.~\cite{ma_robot_2026} organize human-video-based learning into task-, observation-, and action-oriented transfer pathways. Feng et al.~\cite{feng_human_2026} distinguish latent actions, predictive world models, and explicit 2D/3D supervision for VLA learning. Our review examines how EHV observation conditions affect robot manipulation through Ego2Robot learning. We therefore compare what information is derived, how robots use it during training or deployment, and what additional data or grounding is required.

Figure~\ref{fig:taxonomy} illustrates the four learning paradigms and the data resources that support them. We distinguish four paradigms according to how human video contributes to Ego2Robot learning: 1)~\textbf{explicit embodiment transfer} explicitly converts EHV's human manipulation behavior into robot-format training data; 2)~\textbf{explicit action and task supervision} extracts interpretable signals from EHV that guide policy learning or execution; 3)~\textbf{egocentric video representation pretraining} learns visual representations from EHV for downstream manipulation; and 4)~\textbf{latent action and dynamics modeling} learns action abstractions and predictive models from EHV's visual experience. Alongside them, we consider EHV as the primary source of human experience, robot demonstrations as a source of embodiment-specific grounding, and synthetic and simulated data as resources for augmentation and physically grounded learning.

The contributions of this paper are threefold. First, we organize Ego2Robot learning into four paradigms and review their supporting data sources, as shown in Figure~\ref{fig:taxonomy}. Second, we compare the information used by these paradigms and examine how first-person observation constraints affect their additional supervision and grounding requirements. Third, we identify open challenges in physically feasible transfer, agent coordination, in-context adaptation, controlled evaluation, and predictive control. We consider both training-time knowledge transfer and deployment-time use of demonstrations and feedback.

\begin{table*}[t]
\centering
\caption{Comparison of survey papers related to robot manipulation from video. \emph{Primary scope} identifies each review's central topic and data focus. \emph{Organizing and analytical perspective} summarizes its classification framework and main analytical concerns. Our paper centers on egocentric human video (EHV) with Ego2Robot learning, where robot demonstrations and synthetic/simulated data are considered as complementary resources for grounding and augmentation. VLA: vision-language-action.}
\label{tab:survey_comparison}

\begingroup
\tablestyle{4pt}{1.20}
\renewcommand{\tabularxcolumn}[1]{m{#1}}

\begin{tabularx}{0.98\linewidth}{@{}
M{0.17\linewidth}
C{7mm}
M{0.23\linewidth}
Y@{}}
\toprule
\textbf{Review} &
\textbf{Year} &
\textbf{Primary scope} &
\textbf{Organizing and analytical perspective} \\
\midrule

Eze and Crick~\cite{eze_learning_2025} &
2025 &
Video-based robot manipulation &
Perception foundations and learning approaches for manipulation from passive video \\

McCarthy et al.~\cite{mccarthy_towards_2025} &
2025 &
Generalist robot learning from internet video &
Knowledge extraction, video-to-robot adaptation, and video-informed learning \\

Zheng et al.~\cite{zheng_video_2026} &
2026 &
Manipulation from action-unlabeled video &
Video-to-control interfaces; loop closure, verification, and failure modes \\

Ma et al.~\cite{ma_robot_2026} &
2026 &
Robot learning from general human videos &
Task-, observation-, and action-oriented transfer; couplings with data and learning paradigms \\

Feng et al.~\cite{feng_human_2026} &
2026 &
Human-video-based VLA learning &
Latent actions, world models, and explicit 2D/3D signals; representation bridges and grounding \\

Ye et al.~\cite{ye_data_2026} &
2026 &
Data ecosystems for embodied manipulation &
Five-source data pyramid; scalability, robot alignment, and data-mixing recipes \\

\midrule
\textbf{Ours} &
2026 &
\textbf{Ego2Robot learning} &
\textbf{What information EHV provides, how robots use it, and what additional data or grounding is required} \\

\bottomrule
\end{tabularx}
\endgroup
\end{table*}

\textbf{Scope and literature selection.}
The core literature reviewed in this paper satisfies two criteria: it uses EHV as a source of training data, supervision, or task information, and it connects that information to robot manipulation through policy learning, planning, or execution. Studies using mixed data sources are included when their use of EHV is explicitly documented; their inclusion does not imply that reported gains can be attributed to EHV alone. We consider both peer-reviewed publications and publicly available preprints as literature in this paper. Foundational video-learning methods, perception models, robot demonstrations, and synthetic data resources are discussed as methodological context or complementary grounding rather than evidence of Ego2Robot learning in their own right.

The remainder of this paper is organized as follows.
Section~\ref{sec_main_1} presents the problem formulation and four learning paradigms.
Section~\ref{sec_main_2} reviews the supporting training data landscape.
Section~\ref{sec_challenges} discusses open challenges and future directions, and Section~\ref{sec_conclusion} concludes the paper.


\section{Ego2Robot Learning Paradigms}\label{sec_main_1}

\textbf{Problem formulation.}
Given egocentric human videos $\mathcal{D}_{\mathrm{h}}$, Ego2Robot learning aims to acquire or adapt robot manipulation behavior from observed human interactions, optionally with complementary robot data $\mathcal{D}_{\mathrm{r}}$. Each human video is a frame sequence $V_{\mathrm{h}} = \{I_{\tau}^{\mathrm{h}}\}_{\tau=1}^{T_{\mathrm{h}}}$. A robot policy $\pi_\theta(a_t \mid o_{\leq t}, g, c_t)$ predicts an action $a_t$ from visual and proprioceptive observation history $o_{\leq t}$, an optional task specification $g$, and optional deployment-time context $c_t$. The task specification may be a language instruction or goal image; context may contain a demonstration, retrieved experience, or previous robot observation-action pairs and feedback. EHV can shape the policy parameters during training or provide task-relevant context during deployment. These uses are complementary and do not eliminate the need to ground human observations in the target robot's capabilities.

Following the taxonomy in Figure~\ref{fig:taxonomy}, we examine what manipulation-relevant information is derived from EHV, how robots use it, and what additional grounding or assumptions are required. Training-time transfer and deployment-time conditioning specify when EHV is used.

\subsection{Explicit Embodiment Transfer}\label{sec_main_1_1}
Explicit embodiment transfer converts egocentric human video into robot-format action trajectories or visual observations for policy learning. We review three components of this process: hand pose and HOI reconstruction, motion retargeting, and Ego2Robot video translation.

\subsubsection{Hand Pose and HOI Reconstruction}\label{sec_taxonomy_hand_pose}
\begin{figure}[htbp]
\centering
\includegraphics[width=\textwidth]{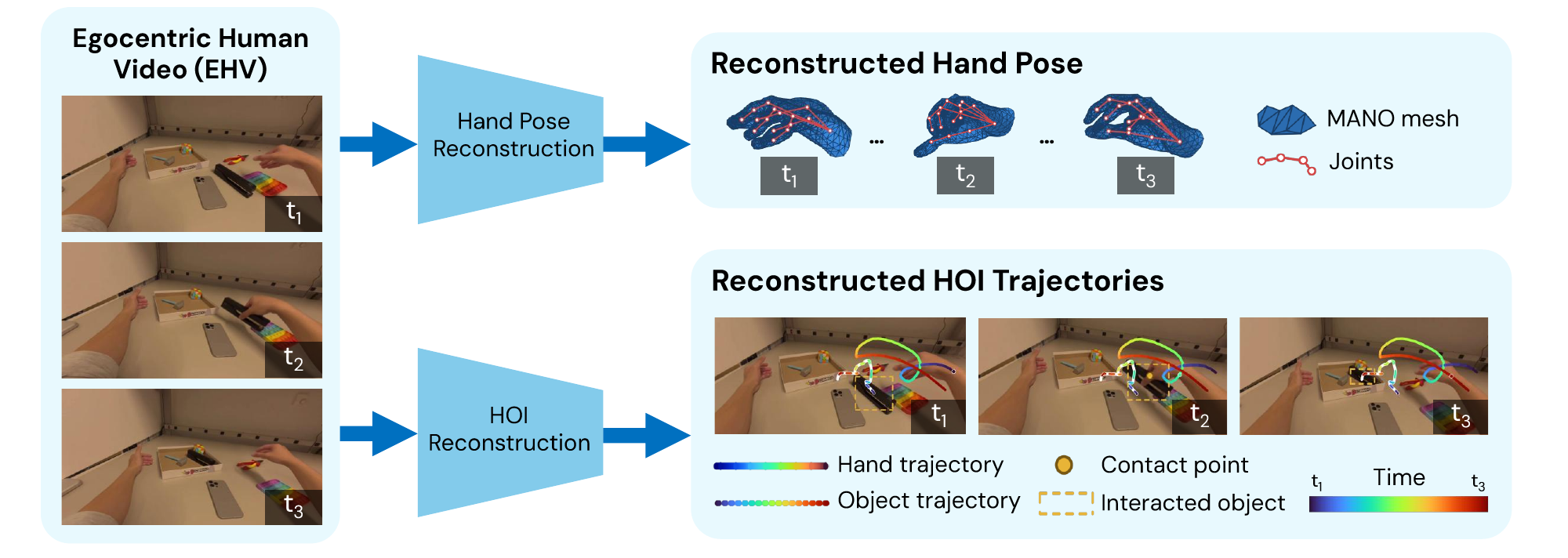}
\caption{Hand pose and HOI reconstruction from EHV. It recovers hand meshes and joints, interaction trajectories, and contact locations for subsequent retargeting. Trajectory colors indicate temporal progression; yellow boxes and markers identify manipulated objects and contacts. Frames from EgoDex~\cite{hoque_egodex_2026}; meshes and overlays are schematic.}
\label{fig:sec_main_1_1_fig}
\end{figure}

The target of hand reconstruction is to estimate hand keypoints, hand meshes~\cite{zhu_mass_2026}, and hand-object trajectories from EHV, as shown in Figure~\ref{fig:sec_main_1_1_fig}. Motion retargeting (Sec. \ref{sec_taxonomy_motion_retargeting}) uses these trajectories to derive robot action labels for imitation learning. We separate methods that reconstruct the hands only from those that additionally recover object motion and hand-object relationships.

\textbf{Hand pose reconstruction.}
Hand pose reconstruction estimates hand joint positions and meshes from monocular images or video frames.
Rong et al.~\cite{rong_frankmocap_2021} introduced FrankMocap for whole-body reconstruction by integrating independently regressed body and hand poses. HaMeR~\cite{pavlakos_reconstructing_2024} improves transformer-based hand reconstruction by increasing model capacity and the diversity of annotated training data. Recently, WiLoR~\cite{potamias_wilor_2025} combines hand localization with mesh reconstruction and demonstrates smooth video tracking without an explicit temporal module. Camera-frame hand estimates do not separate hand motion from ego-motion or provide a consistent world-space trajectory.

Moreover, HaWoR~\cite{zhang_hawor_2025} addresses the moving-camera setting by combining camera-space hand reconstruction with world-space camera trajectory estimation and infilling hand motion when the hands leave the field of view. EgoHandICL~\cite{xie_egohandicl_2026} uses exemplar retrieval guided by a vision-language model (VLM) for in-context hand reconstruction, and Hand-4DGS~\cite{bae_hand-4dgs_2026} combines mesh-guided Gaussian representations with temporal convolutions to reconstruct dynamic hands. EventEgoHands~\cite{hara_eventegohands_2025} and EgoEV-HandPose~\cite{wang_egoev-handpose_2026} process event streams to address motion blur and low-light conditions. These approaches require event-camera recordings and cannot directly process existing RGB-only video collections.

\textbf{HOI reconstruction.}
HOI reconstruction jointly estimates hand and object motion, together with their spatial relationships and contact states, from EHV data. In particular, WHOLE~\cite{ye_whole_2026} jointly reconstructs world-space hand and object motion with a generative interaction prior, given object templates. EgoGrasp~\cite{fu_egograsp_2026} reconstructs world-space HOI for open-vocabulary objects using body-guided diffusion for hand estimation and an HOI prior for object-pose infilling. EgoInfinity~\cite{wang_egoinfinity_2026} integrates reconstruction, metric calibration, and interaction-aware refinement into a video-to-action pipeline, producing hand trajectories, object poses, and contact-relevant states for subsequent retargeting. Object-motion and contact estimates provide retargeting constraints that hand trajectories alone cannot supply.

\textbf{Discussion.}
Camera-frame pose accuracy is insufficient for motion retargeting when trajectories drift, coordinate frames disagree, or reconstructed hand-object relations are inconsistent. In addition, occlusion, monocular depth ambiguity, and camera tracking errors can distort trajectories or introduce drift, potentially corrupting the robot action labels derived from them. World-space tracking and joint hand-object reconstruction address camera-motion ambiguity and interaction inconsistency, respectively~\cite{zhang_hawor_2025, ye_whole_2026}.

\subsubsection{Motion Retargeting}\label{sec_taxonomy_motion_retargeting}
\begin{figure}[htbp]
\centering
\includegraphics[width=\textwidth]{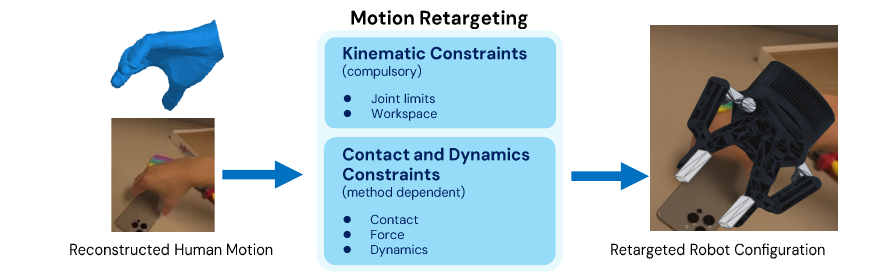}
\caption{Motion retargeting transfers the human motion recovered from EHVs onto the target robot embodiment subject to its joint limits and workspace constraints. Additional contact and dynamics constraints are method-dependent. The mapping is schematic, illustrated with a human grasp frame from EgoDex~\cite{hoque_egodex_2026}. The robot overlay is an illustrative composite.}
\label{fig:sec_main_motionretargeting}
\end{figure}

\looseness=-1
Motion retargeting maps reconstructed hand motion (Sec. \ref{sec_taxonomy_hand_pose}) to robot configurations, seeking to preserve hand-object alignment and contact structure, as illustrated in Figure~\ref{fig:sec_main_motionretargeting}. Its outputs can serve as reference trajectories for control or as demonstrations for policy learning. The demonstrated motion must be adapted to the target robot's kinematic and physical limits.

\textbf{Kinematic retargeting.}
Kinematic methods establish correspondences between human motion and robot configurations through geometric objectives and embodiment-specific constraints. Bharadhwaj et al.~\cite{bharadhwaj_zero-shot_2023} map predicted human hand trajectories to robot motions for zero-shot manipulation without in-domain robot trajectories. Related work on captured hand motion provides complementary tools: DexCap~\cite{wang_dexcap_2024} combines wearable motion capture with retargeting for dexterous imitation learning, and Smooth Operator~\cite{malate_smooth_2026} uses gradient-free sampling to reduce jitter in real-time hand retargeting. These methods provide motion targets, but geometric correspondence alone does not establish whether the robot can maintain the contacts needed to complete the task.

\textbf{Contact- and physics-aware retargeting.}
Interaction-aware methods supplement motion matching with constraints on the hand-object relationship. Wu et al.~\cite{wu_toporetarget_2026} proposed TopoRetarget to preserve interaction structure through sparse graphs over hand and object keypoints. ObjRetarget~\cite{lai_objretarget_2026} combines arm constraints with object-aware contact geometry. C2Dex~\cite{ren_c2dex_2026} uses stable object-side contacts as shared targets for reconstruction and retargeting, followed by residual reinforcement learning to refine the motion in simulation. Beyond geometric contact preservation, SPIDER~\cite{pan_spider_2026} uses physics-based sampling and contact guidance to convert kinematic demonstrations into dynamically feasible robot trajectories. These approaches differ in whether they preserve interaction geometry or additionally account for the dynamics of execution.

\textbf{Use in policy learning and execution.}
Retargeted motion can provide demonstrations for imitation learning, reference targets for reinforcement learning, or objectives for motion planning. Do as I Do~\cite{paliwal_as_2026} converts reconstructed human motion into dexterous robot action sequences. VideoManip~\cite{chen_dexterous_2026} combines contact optimization with demonstration synthesis for policy training. DexImit~\cite{mu_deximit_2026} reconstructs HOI from monocular videos, decomposes the task into subtasks, and schedules the two hands to synthesize interaction-consistent robot trajectories. ViViDex~\cite{chen_vividex_2025} instead uses trajectory-guided rewards to train state-based policies before learning a vision-based policy. Object-centric approaches instead target object motion. Robot See Robot Do~\cite{kerr_robot_nodate} reconstructs articulated object motion from a monocular demonstration and a static multiview object scan, then plans bimanual robot motions to reproduce the object-part trajectories. DexMachina~\cite{mandi_dexmachina_2025} learns policies that track demonstrated object states, allowing robot-specific motions rather than requiring direct imitation of human joint trajectories.

\textbf{Discussion.}
Motion retargeting trades fidelity to the human demonstration against feasibility for the target robot. Errors in reconstructed motion and contact estimates propagate into robot reference trajectories, and embodiment differences can make direct motion matching infeasible. A trajectory that satisfies joint limits can still lose contact or slip because forces and friction are unobserved. Contact and dynamics constraints narrow the feasible trajectories, but errors in reconstructed contacts or the assumed physics still affect execution.

\subsubsection{Ego2Robot Video Translation}\label{sec_taxonomy_video_translation}
\begin{figure}[htbp]
\centering
\includegraphics[width=\textwidth]{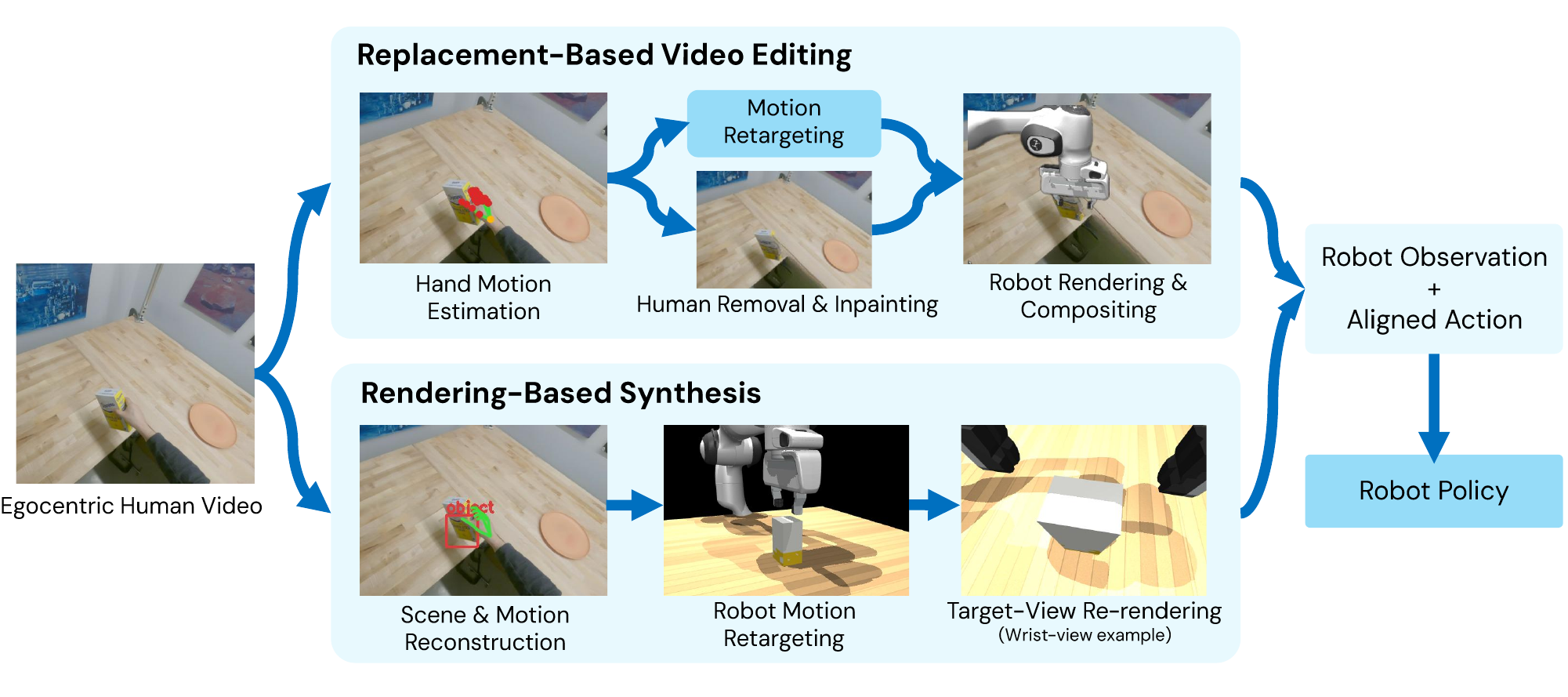}
\caption{Two routes for Ego2Robot video translation: replacement-based editing replaces the human embodiment within the retained scene; rendering-based synthesis reconstructs and re-renders interactions from a target viewpoint. Both video translations produce robot observations with aligned actions for policy learning. Example images from WARPED~\cite{freeman_warped_2026}.}
\label{fig:sec_main_1_1_fig2}
\end{figure}

Ego2Robot video translation synthesizes robot-format observations from the hand-object motion, object states, and scene appearance available in EHV. Motion retargeting converts (Sec. \ref{sec_taxonomy_motion_retargeting}) human motion into robot action trajectories. Video translation instead converts human video into visual observations that match the target robot embodiment and camera viewpoint. Paired with retargeted actions, these observations support Ego2Robot learning. It has two routes: replacement-based video editing and rendering-based synthesis, as shown in Figure~\ref{fig:sec_main_1_1_fig2}.

\textbf{Replacement-based video editing.}
These methods replace the human embodiment with a rendered robot and retain most of the original scene. Phantom~\cite{lepert_phantom_2026} estimates hand poses, removes the human arm by inpainting, and overlays a rendered robot to form robot-compatible observation-action pairs. Masquerade~\cite{lepert_masquerade_2025} applies this approach to in-the-wild egocentric videos, overlaying a bimanual robot that follows recovered hand trajectories. EgoEngine~\cite{liu_egoengine_2026} generates dexterous robot observation videos paired with task-aligned action trajectories under feasibility constraints.

\textbf{Rendering-based synthesis.}
Scene reconstruction allows observations to be rendered from the target robot viewpoint. WARPED~\cite{freeman_warped_2026} tracks hand-object motion from monocular EHV, retargets the trajectories, and uses Gaussian splatting to generate wrist-view observations for policy learning. AgenticFocus~\cite{kolomiets_agenticfocus_2026} restores occluded object geometry and pairs mixed-reality humanoid observations with synchronized robot actions and states. RoboPaint~\cite{fan_robopaint_2026} renders tactile-aware retargeted trajectories in simulation, but relies on multi-camera, glove, and tactile capture rather than monocular EHV alone.

\textbf{Use in policy learning.}
Manipulation policies either train on translated demonstrations directly or use them for pretraining before adaptation on robot data. Phantom and EgoEngine report deployment without robot demonstrations in their evaluated settings~\cite{lepert_phantom_2026, liu_egoengine_2026}. Masquerade~\cite{lepert_masquerade_2025} instead pretrains a visual encoder through future robot-keypoint prediction and retains this auxiliary objective during policy fine-tuning on robot demonstrations. At a larger scale, Wang et al.~\cite{wang_ego2robot_2026} combine action retargeting, visual synthesis, and quality curation to generate data for joint VLA pretraining with robot demonstrations.

\textbf{Discussion.}
The synthesized images need to remain consistent with the retargeted actions. Inpainting can alter manipulated objects or contact regions, and reconstruction or retargeting errors can distort the robot-object configuration. Rendering allows the camera viewpoint to be adjusted, but it inherits errors from scene reconstruction and may produce images that differ from those encountered during deployment.

\subsubsection{Summary}
Explicit embodiment transfer combines reconstructed hand-object geometry, retargeted motion, and synthesized robot observations to produce robot-format supervision. The resulting data can support direct policy learning or pretraining followed by adaptation with robot data.

Methods that do not use robot demonstrations may still require additional scene data. WARPED, for example, uses a separate workspace scan and geometric alignment~\cite{freeman_warped_2026}.
Reports should state the fraction of usable trajectories after conversion, the conversion cost, and the calibration, scene or object reconstruction, simulation, and robot data required. We distinguish methods that directly connect EHV to robot behavior from general-video, perception-only, and robot-data methods used to explain related mechanisms.

\subsection{Explicit Action and Task Supervision}\label{sec_main_1_2}
Methods in this paradigm extract action representations, affordances, and task-progress or reward estimates from EHV as interpretable supervision. Ego2Robot learning uses these selected signals without requiring a complete conversion of the human demonstration into robot observations and actions. For example, a recovered trajectory may serve as an intermediate prediction target rather than being directly retargeted into robot commands. 

The three supervision types serve different purposes: 1) \textbf{Explicit intermediate action representations} describe how manipulation should proceed; 2) \textbf{affordance representation} encodes opportunities for interaction conditioned on the task and embodiment; and 3) \textbf{task-progress and reward inference} provides signals for assessing progress or guiding behavior toward task completion. These signals can supervise action prediction, guide interaction selection, or define objectives for policy optimization and planning. Across the three categories, we examine the reliability of the extracted supervision and its applicability to robots.

\subsubsection{Explicit Intermediate Action Representations}\label{sec_taxonomy_intermediate_actions}
Explicit intermediate action representations express manipulation through identifiable motion variables or interaction targets that connect EHV to Ego2Robot learning and execution. Rather than directly prescribing robot joint commands, they provide targets for prediction or planning with embodiment-specific control handled downstream. We organize these interfaces into trajectory- and keypoint-based representations, object-centric interaction representations, and human-centric and shared action spaces. These groups may overlap, but emphasize different choices about which aspects of the demonstrated behavior to preserve.

\textbf{Trajectory- and keypoint-based representations.}
These methods describe manipulation through the motion of selected image points or hand and end-effector waypoints. Motion Tracks~\cite{ren_motion_2025} uses short-horizon 2D motion tracks as a shared action representation for human and robot demonstrations. A policy predicts these tracks from images, and predictions from two camera views are combined to recover robot end-effector motion. ATM~\cite{wen2024anypointtrajectorymodelingpolicy} predicts language-conditioned trajectories of arbitrary 2D image points from action-free videos and uses these predictions to guide policies learned from limited action-labeled robot demonstrations. Track2Act~\cite{bharadhwaj2024track2actpredictingpointtracks} learns goal-image-conditioned 2D point tracks from human and robot videos. It combines the predicted tracks with initial depth to infer rigid object transforms and robot end-effector trajectories, and then corrects execution with a residual policy trained on robot demonstrations. Uni-Hand~\cite{ma_unihand_2026} predicts 2D and 3D hand waypoints from egocentric inputs, together with hand-object contact and separation states, enabling predicted trajectories to guide robot motion and providing cues for grasp timing. MotionTrans~\cite{yuan_motiontrans_2025} instead uses VR-recorded hand motion, converts it into robot-compatible training data, and applies weighted human-robot co-training. These interfaces require accurate tracking, interaction timing, coordinate alignment, and a mapping from predicted motion to feasible robot actions.

\textbf{Object-centric interaction representations.}
Object-centric interfaces describe manipulation through task-relevant object relationships and motion, reducing dependence on the demonstrator's anatomy. Related RGB-D approaches illustrate this principle: FUNCTO~\cite{tang_functo_2025} uses functional keypoints to establish tool correspondences and plan manipulation from a fixed-camera human demonstration. Human2Any~\cite{cheng_human2any_2026} learns object-object motion priors and combines them with robot-specific feasibility reasoning and planning. These RGB-D methods provide background for the EHV setting. HumanEgo~\cite{wang_humanego_2026} combines entity-relative spatial observations extracted from EHV with explicit hand-pose and grasp targets, using future object motion and projected trajectories as auxiliary supervision. Its interaction-centric tokens describe observations instead of action tokens. Transferring these object-level targets still requires reliable object tracking and robot-feasible motion.

\textbf{Human-centric and shared action spaces.}
Human-centric representations use hand and wrist motion as a common interface for learning from human demonstrations. EgoVLA~\cite{yang_egovla_2025} predicts human wrist and hand actions, which are mapped to robot actions through inverse kinematics and retargeting. VITRA~\cite{li_scalable_2025} segments in-the-wild EHV into hand activities annotated with language descriptions and 3D motion to construct action-labeled pretraining data. Being-H0~\cite{luo_being-h0_2025} uses part-level motion tokenization to encode hand trajectories as prediction targets for VLA pretraining.

Moreover, shared action spaces further make human and robot supervision comparable across embodiments. HAT~\cite{qiu_humanoid_2025} and In-N-On~\cite{cai_-n-_2025} organize human and humanoid data within unified human-centric state-action spaces. EgoMimic~\cite{kareer_egomimic_2024} aligns human hand and robot end-effector trajectories in camera coordinates for joint policy training, with robot data supervising executable joint actions. MiVLA~\cite{yin_mivla_2025} uses kinematic mappings for bidirectional human-robot action alignment and mutual imitation, and Being-H0.5~\cite{luo_being-h05_2026} maps heterogeneous controls into semantically aligned action slots. ACE-Ego-0~\cite{li_ace-ego-0_2026} combines camera-space actions, morphology conditioning, and temporal alignment with reliability-aware weighting of human-derived pseudo-actions. For dexterous hands, UniDex~\cite{zhang_unidex_2026} maps functionally similar actuators to shared coordinates. Joint training still requires consistent coordinate and timing conventions and embodiment-specific mappings from shared actions to robot commands. Recently, Qwen-RobotManip~\cite{yuan2026qwenrobotmaniptechnicalreportalignment} combines a canonical state-action representation with camera-frame end-effector motions and synthesized robot demonstrations from egocentric human data. Zeva-Ego~\cite{huang2026zevaegoegocentricmidtrainingincontext} uses camera-frame action targets and subtask supervision for EHV-based mid-training. These are explicit supervision mechanisms; their deployment-time use of interaction history is a separate adaptation mechanism discussed in Section~\ref{sec_taxonomy_in_context}.

\textbf{Discussion.}
Explicit action representations must balance task-relevant precision with cross-embodiment applicability. Hand trajectories preserve motion detail but require robot-specific interpretation. Object-centric targets leave the robot's realization of the interaction underdetermined. Shared action spaces make heterogeneous supervision comparable without guaranteeing that corresponding actions have equivalent physical effects. They also do not eliminate visual domain differences: EgoBridge~\cite{punamiya_egobridge_2025} addresses this complementary issue by aligning joint distributions of policy features and actions.

\subsubsection{\texorpdfstring{Affordance Representation}{Affordance Representation}}\label{sec_taxonomy_affordance}
\begin{figure}[htbp]
\centering
\includegraphics[width=\textwidth]{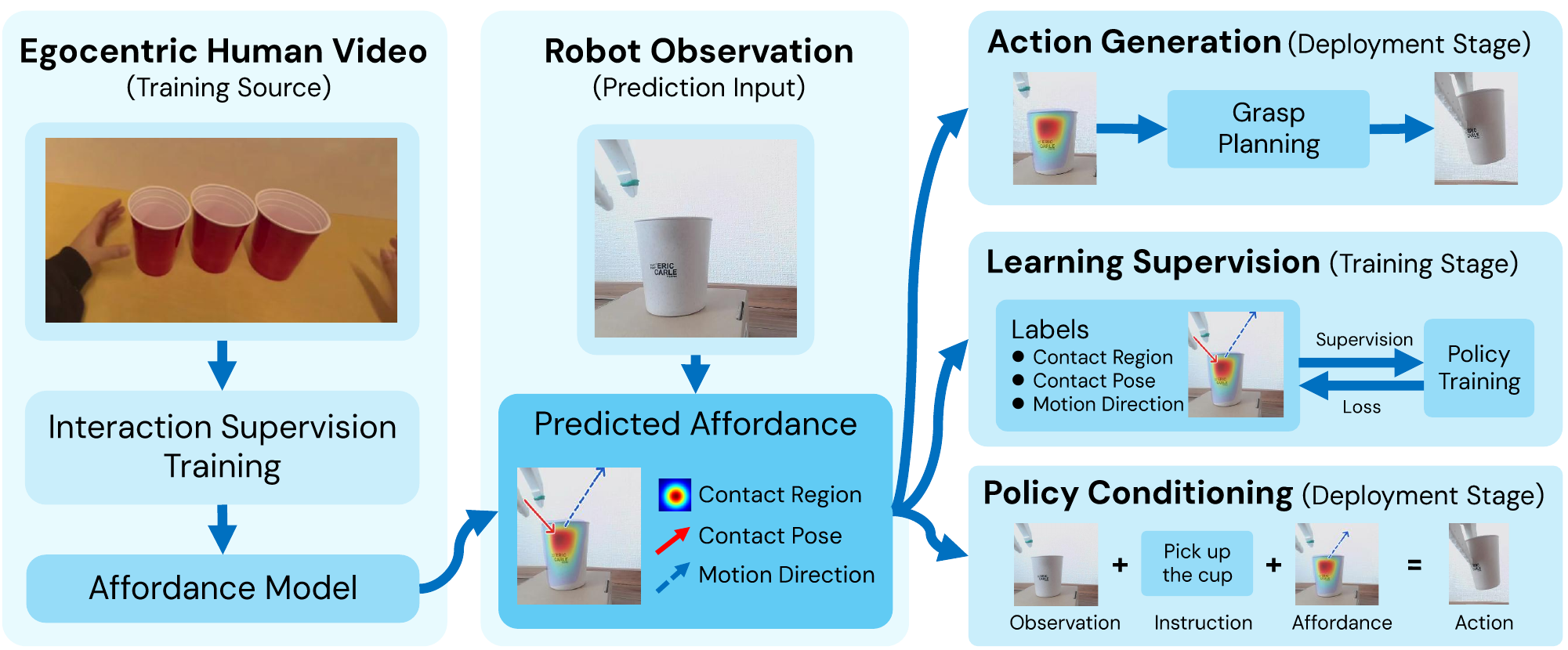}
\caption{Human-derived affordances and their three complementary uses: action generation, learning supervision, and policy conditioning. Images from EgoDex~\cite{hoque_egodex_2026} and so100\_grasp\_cup~\cite{so100_grasp_cup}.}
\label{fig:affordance_roles}
\end{figure}

Affordance representations encode task-relevant interaction opportunities inferred from EHV, including contact regions, interaction poses, and possible post-contact motion. Affordances link possible actions to the current object and scene~\cite{yao_hammer_2026}. Which interactions are available depends on the task, scene state, and robot embodiment. As illustrated in Figure~\ref{fig:affordance_roles}, we discuss three complementary uses: grounding affordances for action generation, using affordances as learning supervision, and conditioning policies and exploration on affordance predictions. These roles are not mutually exclusive.

\textbf{Affordance grounding for action generation.}
Predicted affordances can be treated as geometric targets or constraints for grasp synthesis, motion planning, and control. FSAG~\cite{han_fsag_2026} extracts finger-specific contact regions from human demonstrations and combines them with depth-derived geometry to guide grasp refinement. RoboPCA~\cite{xiao_robopca_2026} jointly predicts task-conditioned contact regions and poses, reducing inconsistencies between separately estimated interaction locations and end-effector orientations. VidBot~\cite{chen_vidbot_2025} learns 3D interaction trajectories from monocular human videos, and BridgeACT~\cite{han_bridgeact_2026} predicts task-relevant grasp regions and 3D motion affordances, using a grasping module and closed-loop controller to execute them. These approaches connect interaction predictions to execution, but accurate affordance grounding alone does not ensure the feasibility of resulting motion for the target robot.

\textbf{Affordance-based learning supervision.}
Human-derived affordances can influence learned representations or policies without acting as explicit targets at every execution step. HRP~\cite{srirama_hrp_2024} extracts hand, object, and contact cues from human video and uses them to adapt pretrained visual encoders for Ego2Robot learning. DexMani~\cite{chen_dexmani_2026} further transfers contact-conditioned manipulability evolution, i.e., a prior describing how contact transitions change the object-rotation directions available to a hand. This prior guides reinforcement learning across different dexterous hands. The human-derived cues supervise learning instead of specifying complete robot commands, so extraction errors and embodiment mismatch can bias the resulting policy.

\textbf{Affordance-conditioned policies and exploration.}
Predicted affordances can condition policies, specify goals, or focus exploration on interaction regions. VRB~\cite{bahl_affordances_2023} learns contact locations and post-contact motion from human videos and uses these predictions across imitation learning, exploration, goal-conditioned learning, and reinforcement-learning action parameterization. Visual Affordance Prediction~\cite{bharadhwaj_visual_2023} models possible future visual states from passive interaction data, providing goal proposals for exploration and goal-conditioned learning. VAPO~\cite{borja-diaz_affordance_2022} and HULC++~\cite{mees_grounding_2023} use affordances learned from robot play data to guide goal reaching, and Affordance2Action~\cite{liu_affordance2action_2026} provides task-conditioned part grounding that supports affordance-conditioned policies. These examples differ in their supervision sources and should not all be interpreted as learning directly from EHV.

\textbf{Discussion.}
Affordances inferred from human demonstrations describe observed interaction possibilities where an exhaustive set of valid robot actions cannot be found. Demonstrated contacts may not suit the target robot's morphology, and correctly localized regions do not determine the forces or motion required for successful execution. Note that target errors can misguide planning, noisy supervision can bias learned representations, and incorrect policy conditioning can direct exploration toward unsuitable interactions.

\subsubsection{Task-Progress and Reward Inference}\label{sec_taxonomy_task_reward}
\begin{figure}[htbp]
\centering
\includegraphics[width=\textwidth]{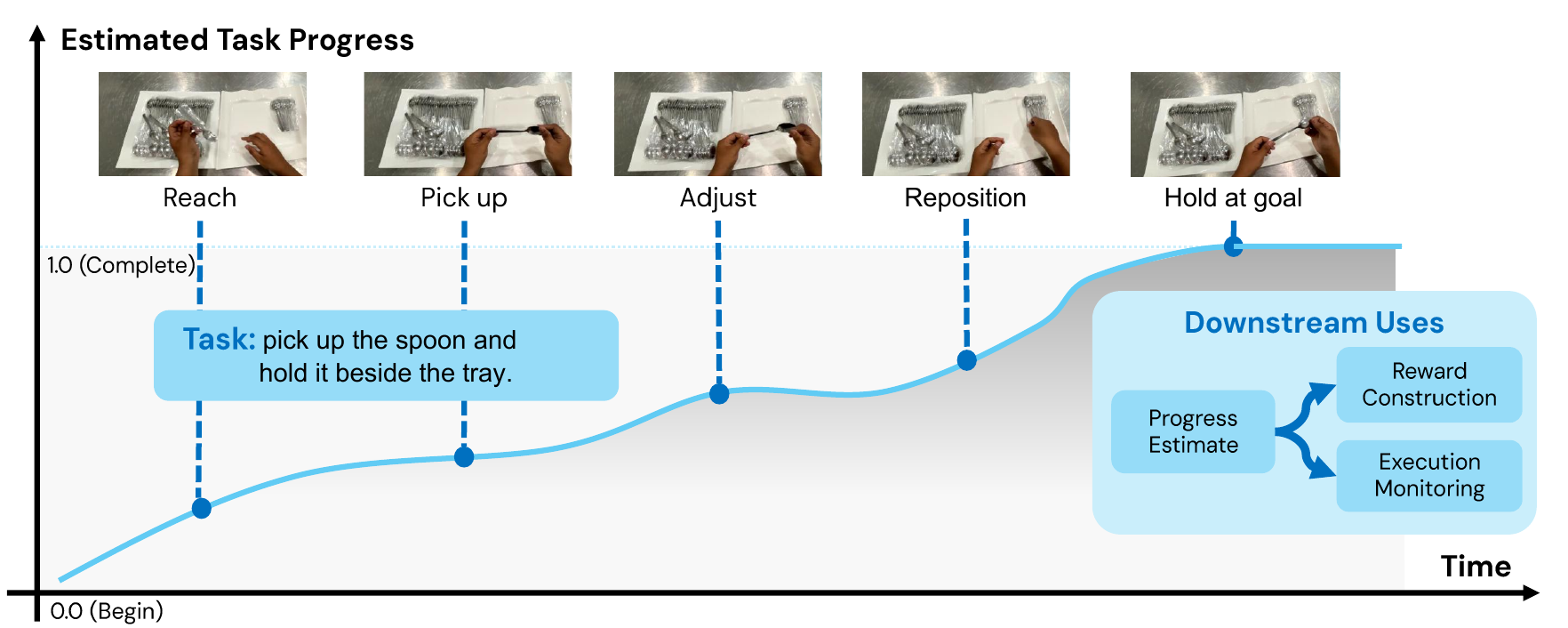}
\caption{An illustrative progress estimate for picking up a spoon and holding it beside the tray. Progress ranges from 0 at task initiation to 1 when the spoon is held at the goal. Progress is useful information to support reward construction and execution monitoring. Video frames from EgoVerse~\cite{punamiya_egoverse_2026}.}
\label{fig:task_progress_reward}
\end{figure}

\looseness=-1
Task-progress and reward inference derive feedback from EHV for policy optimization, planning, and execution monitoring in Ego2Robot learning. Progress estimates track how much of a task has been completed, rewards guide optimization, and values estimate expected future return or goal reachability. EHV can supply supervision without recorded robot actions, although downstream learning may still require robot data or interaction. Progress is estimated over a manipulation sequence, as shown in Fig.~\ref{fig:task_progress_reward}. We organize the methods into task-similarity and goal-conditioned rewards, generative- and motion-consistency-based rewards, stage- and event-level extraction, and generalist reward and value learning.

\textbf{Task-similarity and goal-conditioned rewards.}
EHV can define the desired behavior or goal against which robot execution is evaluated. In particular, DVD~\cite{chen_learning_2021} combines human videos with a small amount of robot data to learn whether two videos depict the same task, using this similarity as a reward for visual model predictive control. XIRL~\cite{zakka2021xirlcrossembodimentinversereinforcement} learns temporally aligned representations from cross-embodiment demonstration videos and uses distance to a goal embedding as a reward for robot reinforcement learning. Rank2Reward~\cite{yang2024rank2rewardlearningshapedreward} learns task progress by temporally ranking demonstration frames and converts these estimates into rewards for policy learning. Differently, VIP~\cite{ma_vip_2023} learns a goal-conditioned visual representation from action-free Ego4D videos, with embedding distances implicitly defining a value function from which rewards can be constructed. LIV~\cite{ma_liv_2023} extends value-based representation learning to language and image goals, providing dense feedback for downstream robot control. Their scores evaluate task attainment without prescribing robot actions, but visually similar states across embodiments or scenes may differ in whether the task is complete.

\textbf{Generative- and motion-consistency-based rewards.}
These methods score robot execution using the confidence of a model trained on expert videos or the similarity between observed object motion and predicted or demonstrated trajectories. Diffusion Reward~\cite{huang_diffusion_2024} trains a conditional video diffusion model on expert videos and uses negative conditional entropy to encourage expert-like behavior. Verghese et al.~\cite{verghese_human_2026} learn point-motion prediction from EHV and reward agreement between predicted and observed object motion, using this feedback for on-robot reinforcement learning initialized with robot demonstrations. As a complementary RGB-D approach, X-Sim~\cite{dan_x-sim_2025} reconstructs a simulation from human video and uses demonstrated object trajectories to define rewards for policy learning before sim-to-real transfer. These rewards need no explicit progress labels, but confident predictions or matching motion can still accompany task failure. Model error, tracking error, and valid alternative executions can all change the inferred reward.

\textbf{Stage- and event-level extraction.}
Human demonstrations can provide a temporal structure for evaluating long-horizon execution. AVID~\cite{smith_avid_2020} translates human demonstrations into robot-appearance videos and uses them to support stage-wise reward learning and model-based control. Other methods extract task structure without learning a reward. SeeDo~\cite{wang_vlm_2024} combines keyframe selection with VLM reasoning to produce subtask-level action plans, and EgoLoc~\cite{ma_egoloc_2025} localizes hand-object contact and separation events in egocentric videos. Such outputs can support subgoal specification or stage-aware evaluation, but event timestamps and action plans do not themselves establish task completion or define a reward. Stage-aware evaluation therefore requires boundaries that mark task-relevant state changes.

\textbf{Generalist reward and value learning.}
These models use different supervision targets across heterogeneous trajectory datasets. Liang et al.~\cite{liang_robometer_2026} introduced Robometer, a reward model that combines frame-level progress supervision with trajectory-level preference comparisons and incorporates failed and suboptimal behavior into training. ProcVLM~\cite{feng_procvlm_2026} estimates progress from procedural stages and intra-stage visual changes, reasoning about the atomic actions that remain. RynnValue~\cite{huang_rynnvalue_2026} learns goal-conditioned temporal distance from timestamp-derived labels and converts its estimates into dense rewards through potential-based shaping. Their data sources and evaluation settings should be distinguished from EHV-only learning.

Evaluative signals also support policy refinement and execution monitoring. Large Reward Models~\cite{wu_large_2026} combine progress, completion, and temporal-contrastive rewards to guide online refinement of imitation-learned policies. PRTS~\cite{zhang_prts_2026} incorporates goal-reachability learning into VLA pretraining and does not provide a separate reward model. At a task-specific scale, ValueFormer~\cite{sa_valueformer_2026} learns distinct value and mistake-detection signals from robot rollouts for training-sample weighting and execution monitoring. These results show how evaluative feedback supports robot learning and monitoring, but do not show whether feedback learned from EHV can provide the same benefits.

\textbf{Discussion.}
We claim that task progress is not always reflected by elapsed time, visual similarity, and prediction confidence. For example, a pause adds elapsed time without advancing the task, and recovery may require undoing earlier actions. Stage-aware estimation can reduce these ambiguities, but incorrect stage boundaries (by temporal grounding) or goals may still produce misleading scores. If these scores are used as rewards, a policy may learn to exploit their errors and obtain high scores without completing the task. Evaluation should therefore include failed attempts and recovery to check whether the feedback reflects actual task progress and completion.

\subsubsection{Summary}
Explicit action and task supervision extracts selected learning signals from EHV. Instead, explicit embodiment transfer converts human behavior into robot-format trajectories or observations. Intermediate action representations provide motion targets or shared action interfaces, affordances identify task-relevant interaction opportunities, and progress, reward, and value estimates provide evaluative feedback.

Motion targets must remain interpretable by the target robot, affordances must suit its interaction capabilities, and evaluative signals must distinguish task achievement from misleading visual or temporal cues. Explicit supervision makes these assumptions easier to inspect, but does not guarantee their validity.

\subsection{Egocentric Video Representation Pretraining}\label{sec_main_1_3}
\begin{figure}[htbp]
\centering
\includegraphics[width=\textwidth]{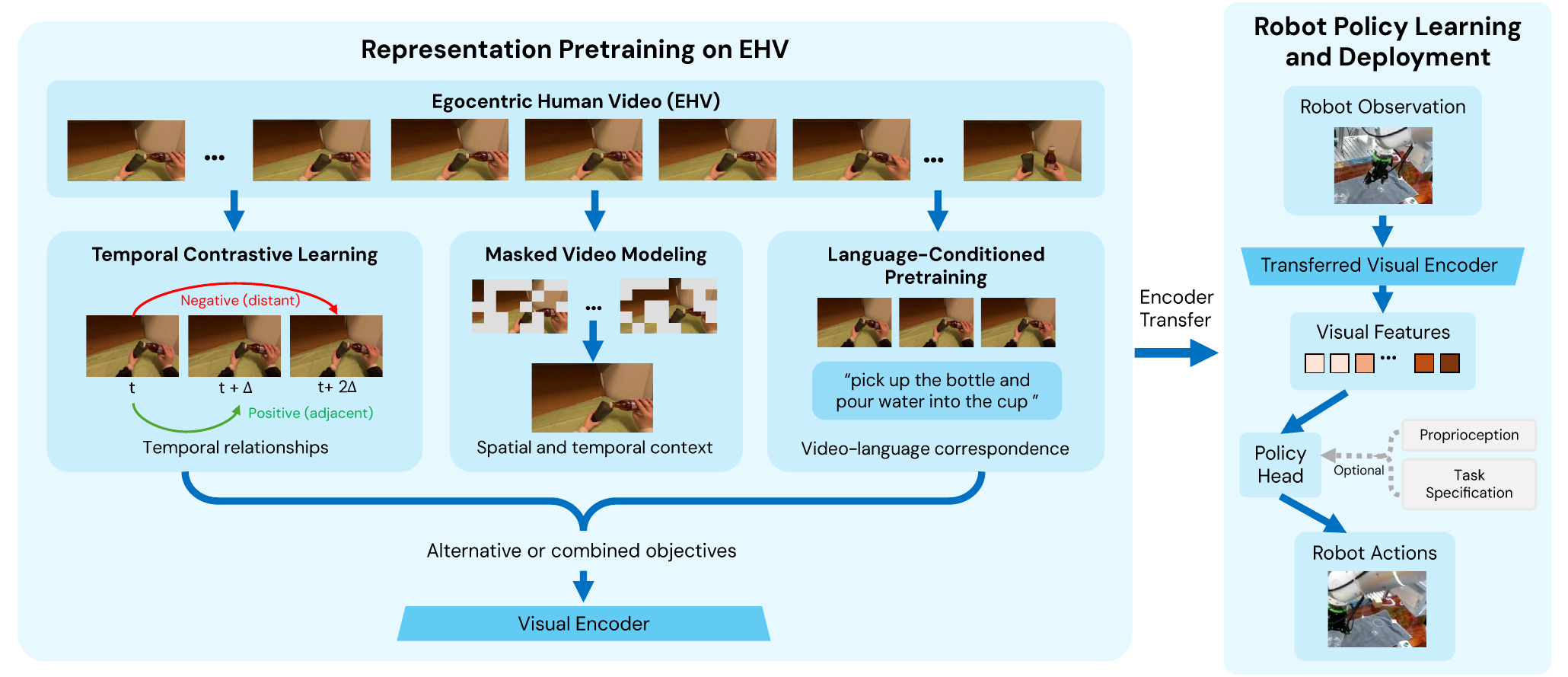}
\caption{Egocentric video representation pretraining through temporal contrastive learning, masked modeling, and language-conditioned objectives. The transferred encoder supplies features for robot policies and may be frozen or fine-tuned. Human and robot frames are from EgoDex~\cite{hoque_egodex_2026} and RH20T~\cite{fang_rh20t_2024}, respectively.}
\label{fig:sec_main_1_1_fig3}
\end{figure}

EHV pretraining trains a visual encoder for downstream robot policies. The encoder may remain frozen or be adapted during policy training. We group the methods by their pretraining signal: temporal frame relationships, masked visual content, or paired language. For each objective, we examine how learning from EHV changes the features supplied to a robot policy. The visual encoder is pretrained with these objectives and transferred to a robot policy, as summarized in Fig.~\ref{fig:sec_main_1_1_fig3}.


\textbf{Related foundation-model distillation.}
Distillation transfers pretrained visual features into encoders used by robot policies. Shang et al.~\cite{shang_theia_2024} proposed Theia to combine multiple vision teachers through feature alignment. X-Distill~\cite{shao_x-distill_2026} transfers DINOv2 features to a compact encoder that can be fine-tuned with a policy. Both methods distill pretrained visual features on ImageNet for downstream robot policy learning.

\subsubsection{Temporal Contrastive Learning}\label{sec_taxonomy_temporal_contrastive}
Temporal contrastive objectives bring the features of selected frame pairs together and separate those of other pairs. For EHV, these objectives should preserve interaction changes and reduce sensitivity to ego-motion and incidental appearance. The resulting encoder supplies observation features for robot policies rather than explicit action targets.

\textbf{Foundational temporal objectives.}
Time-Contrastive Networks~\cite{sermanet_time-contrastive_2018} use synchronized multi-view frames as positive pairs and temporally offset frames as negatives, learning features that are insensitive to viewpoint but sensitive to interaction state. Temporal Cycle-Consistency Learning~\cite{dwibedi_temporal_2019} trains frame embeddings with a cycle-consistency loss to align corresponding moments across videos. The methods below use similar temporal supervision to learn representations of interaction states and task progression from EHV. Besides, they evaluate EHV pretraining for robot manipulation.

\textbf{Representation learning for Ego2Robot policies.}
R3M~\cite{nair_r3m_2022} pretrains a visual encoder on Ego4D~\cite{grauman_ego4d_2022} using temporal contrastive learning, video-language alignment, and sparsity regularization. Its frozen visual features support downstream policies trained from robot demonstrations, without requiring explicit human-action reconstruction. AcTOL~\cite{zhang_provable_2025} pretrains on EPIC-KITCHENS-100~\cite{damen_rescaling_2020} and combines an ordering-aware vision-language contrastive objective with a local Brownian-bridge continuity constraint. This design avoids assuming that later frames always match the instruction better, including after the action has ended. A language-conditioned behavior-cloning policy consumes the pretrained visual and language features. The method was evaluated in simulation and on real robots. 
Ag2x2~\cite{xiong2025ag2x2robustagentagnosticvisual} learns time-contrastive representations from EPIC-KITCHENS~\cite{9084270} frames with the hands removed from the images and their 2D positions retained to preserve bimanual coordination information. Its features define goal-similarity rewards for reinforcement learning. 


\textbf{Discussion.}
Note that temporal distance provides useful supervision, but it does not consistently reflect changes in manipulation state. Nearby frames may straddle a contact transition, and distant frames may show equivalent states during repeated actions or pauses. Ego-motion and occlusion can also change appearance without changing the interaction.

These ambiguities affect the features supplied to robot policies. Strong temporal smoothing can obscure brief contact events, and excessive separation can produce inconsistent features for equivalent states. Policies or goal-similarity rewards may then become sensitive to execution timing instead of task state. Evaluation should therefore examine whether representations preserve relevant state transitions across viewpoints and execution speeds, and whether they improve downstream control.

\subsubsection{Masked Video Modeling}\label{sec_taxonomy_masked_video}
Masked video modeling aims to predict missing visual content from visible spatial or temporal context. Here, we include both masked reconstruction of individual frames sampled from EHV and prediction across frames. These targets encourage scene and interaction features without requiring robot action labels during human-video pretraining. The pretrained encoder subsequently provides visual inputs for Ego2Robot learning.

\begin{figure}[htbp]
\centering
\includegraphics[width=\textwidth]{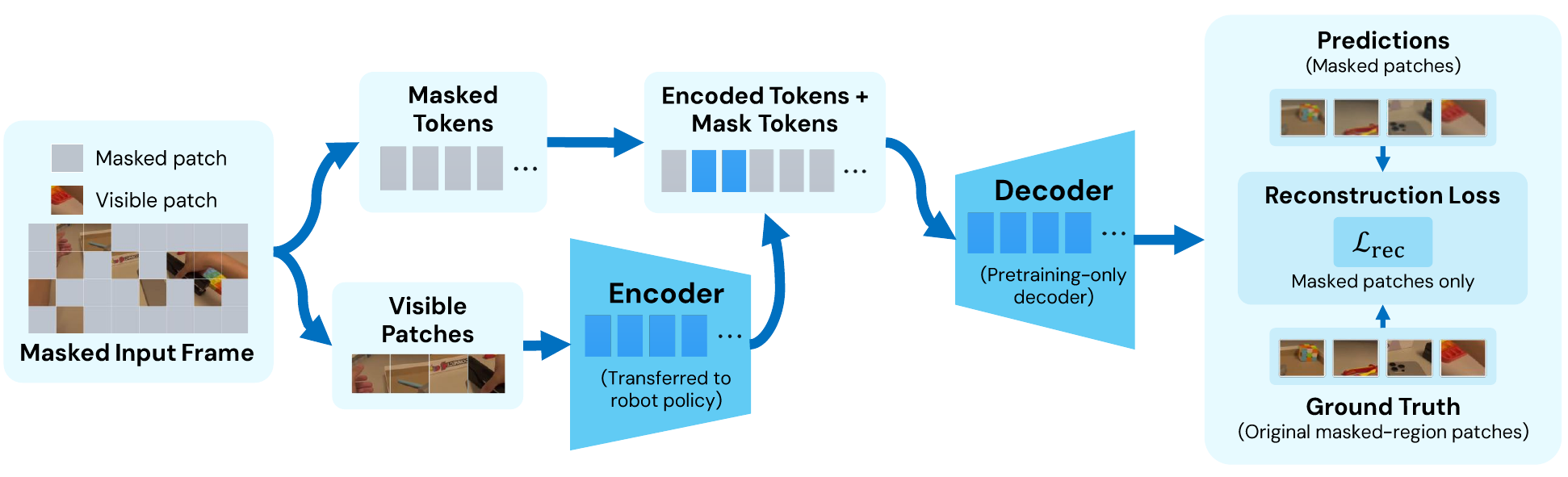}
\caption{Spatial masked pretraining on EHV frames~\cite{he_masked_2021, radosavovic_real-world_2022}. The encoder processes visible patches; encoded features and mask tokens enter the decoder to reconstruct masked content. Predictions are compared with ground-truth patches. The encoder transfers to Ego2Robot learning. The decoder is discarded. Video frames are from EgoDex~\cite{hoque_egodex_2026}.}
\label{fig:masked_video_modeling}
\end{figure}

\textbf{Spatial masked prediction.}
Masked Autoencoders~\cite{he_masked_2021} establish the basic mechanism: an encoder processes visible image patches, and a lightweight decoder reconstructs the missing content. Figure~\ref{fig:masked_video_modeling} illustrates spatial masked pretraining on individual EHV frames and the transfer of the learned encoder to Ego2Robot learning. MVP~\cite{radosavovic_real-world_2022} applies this approach to a mixture of egocentric and other visual data, including Ego4D and EPIC-KITCHENS, and uses the frozen encoder to train policies from robot demonstrations. Its real-world experiments demonstrate transfer across tasks and embodiments. VC-1~\cite{majumdar_where_2023} further examines MAE pretraining at scale, showing that larger and more diverse datasets improve average embodied-task performance without producing a universally optimal encoder; task- or domain-specific adaptation remains important.

\textbf{Spatiotemporal masked prediction.}
Temporal prediction complements frame-wise reconstruction by making changes across observations part of the learning target. As methodological background, MotionMAE~\cite{yang_self-supervised_2022} jointly reconstructs masked video content and predicts temporal differences between nearby frames. For Ego2Robot learning, STP~\cite{yang_spatiotemporal_2024} pretrains on EgoClip clips derived from Ego4D, combining current-frame reconstruction with future-frame prediction through separate decoders. The temporal decoder uses current-frame features and a small set of visible future-frame patches as conditioning. The resulting image encoder supplies frozen features for policies learned from robot demonstrations, with evaluation in simulated and real-world manipulation. Future-frame information is used during pretraining, not required at deployment.

\textbf{Related robot-centric extensions.}
Masked modeling also supports richer observation modalities. 3D-MVP~\cite{qian_3d-mvp_2025} pretrains a multi-view encoder on rendered RGB-D observations of 3D assets from Objaverse. EmbodiedMAE~\cite{dong_embodiedmae_2025} jointly models RGB, depth, and point clouds from DROID-3D robot demonstrations. These studies illustrate how reconstruction objectives can support spatial representations for manipulation, but use synthetic or robot data, so they are background examples rather than EHV-based evidence.

\textbf{Discussion.}
Random masking does not prioritize hands, manipulated objects, or contact regions. Temporal prediction can also capture camera motion alongside interaction changes. Accurate reconstruction therefore does not establish whether the encoder preserves the hand-object relations and state changes needed for control.

\subsubsection{Language-Conditioned Pretraining}\label{sec_taxonomy_language_pretraining}
The target of language-conditioned pretraining is to learn object, interaction, and task representations from language descriptions paired with EHV. Language can supervise visual-text alignment or condition visual reconstruction and prediction, complementing the temporal and masked objectives discussed above. We first review how EHV is aligned with language and then examine how the resulting representations support robot policy learning.

\begin{figure}[htbp]
\centering
\includegraphics[width=\textwidth]{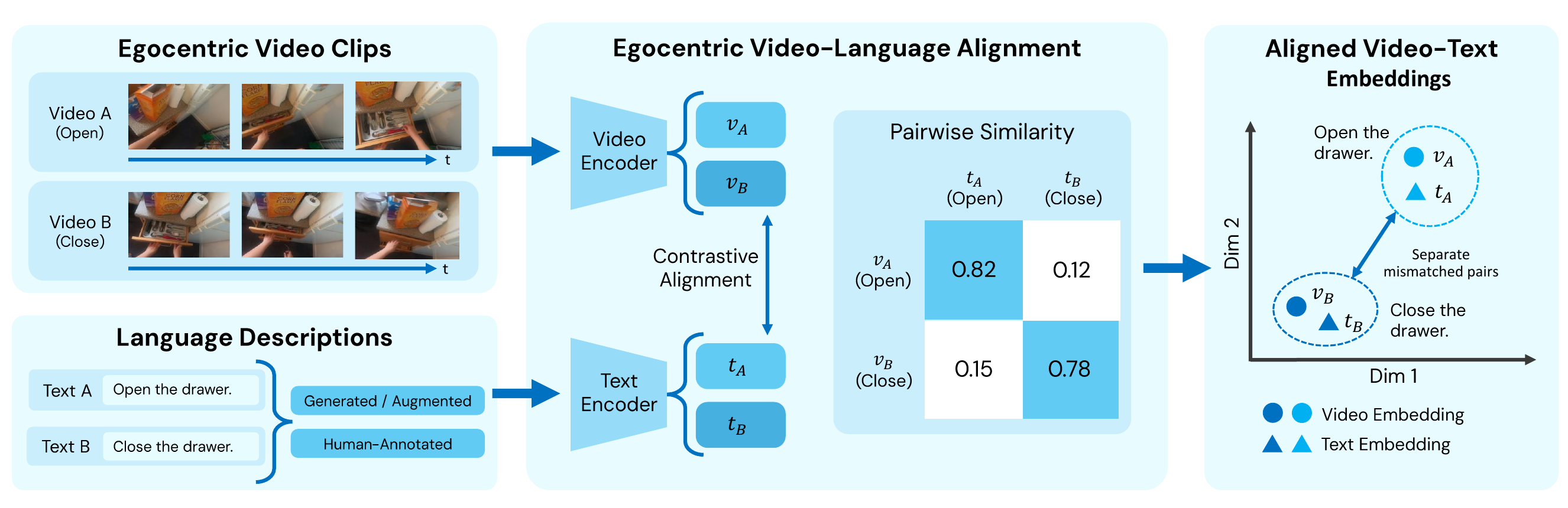}
\caption{Contrastive video-language alignment illustrated with drawer-opening and drawer-closing clips from EPIC-KITCHENS~\cite{damen_rescaling_2020}. Matched video-text pairs are aligned, and mismatched pairs are separated. Text descriptions, similarity scores, and embedding positions are illustrative.}
\label{fig:language_conditioned_pretraining}
\end{figure}

\textbf{Video-language alignment.}
Lin et al.~\cite{lin_egocentric_2022} introduced EgoVLP with the EgoClip dataset, comprising 3.8 million clip-text pairs from Ego4D, and the EgoNCE objective for egocentric video-text alignment. LaViLa~\cite{zhao_learning_2022} adds automatically generated narrations to video-text pretraining. These methods are evaluated on video understanding, not robot control. This contrastive alignment mechanism is shown in Figure~\ref{fig:language_conditioned_pretraining}.

For robot control, LIV~\cite{ma_liv_2023} pretrains on EPIC-KITCHENS using a joint objective that connects visual temporal structure with language and image goals. DecisionNCE~\cite{li_decisionnce_2024} likewise uses EPIC-KITCHENS and learns instruction-grounded representations by treating matched video-language trajectories as implicitly preferred over mismatched pairs. Both use frozen observation and language embeddings as inputs to policies trained through language-conditioned behavior cloning.

\textbf{Language-guided visual prediction.}
Language can condition visual prediction during representation learning. Voltron~\cite{karamcheti_language-driven_2023} combines language-conditioned visual reconstruction with visually grounded language generation, using Something-Something V2 human videos. Using the Ego4D HOI subset, MPI~\cite{zeng_learning_2024} conditions on initial and final keyframes and language to predict an intermediate interaction frame and localize the manipulated object. Its pretraining uses keyframe and object-box annotations. The frozen encoder then supplies features for downstream robot policies. 

\textbf{Fine-grained interaction grounding.}
The representation must distinguish different actions involving similar objects. EgoNCE++~\cite{xu_egocentric_2025} uses verb- and noun-substituted captions as negatives and groups videos sharing nouns as positives. EgoDTM~\cite{xu_egodtm_2025} enriches captions with hand-object cues and combines video-text alignment with pseudo-depth supervision. Their reported gains concern egocentric perception and grounding. Further evaluation on robot tasks is needed to determine whether these gains improve control reliability.

\textbf{Discussion.}
Language descriptions specify task goals but usually omit contact geometry, forces, and detailed execution motions. Coarse narrations or incorrect generated captions can lead models to associate tasks with object identity instead of the intended interaction. For example, recognizing a cup does not distinguish picking it up from pushing it. Evaluation should therefore test whether different instructions in the same scene produce the appropriate changes in robot behavior, beyond measuring video-text retrieval. Joint VLA training on mixed human and robot data, as in Qwen-VLA~\cite{wang_qwen-vla_2026}, must also connect language instructions to executable actions suited to the current scene and robot embodiment.

\subsubsection{Summary}
Egocentric video representation pretraining learns perceptual features from EHV for downstream robot policies. Temporal contrastive learning emphasizes temporal structure, masked modeling captures spatial and temporal context through reconstruction, and language-conditioned objectives highlight interaction and task semantics. The pretrained encoders often remain frozen during policy learning~\cite{nair_r3m_2022, radosavovic_real-world_2022, yang_spatiotemporal_2024}. AcTOL additionally adapts its encoders to the target domain using a small set of robot demonstrations before freezing them for behavior cloning~\cite{zhang_provable_2025}. Despite these different objectives and adaptation strategies, the learned features may emphasize background appearance or coarse task semantics and miss subtle hand-object relations and object state changes needed for control.

\subsection{Latent Action and Dynamics Modeling}\label{sec_main_1_4}
The methods in Section~\ref{sec_main_1_3} learn observation features; this paradigm instead uses EHV transitions as action-level targets or predictive supervision for robot control. We consider two modeling routes. Latent action learning represents visual changes as latent variables that can supervise policy learning or condition robot action generation. World-action modeling predicts future observations or interaction states to support policy training, planning, and action evaluation. The two routes can be combined, with latent actions serving as conditioning signals for predictive models~\cite{gao_dreamdojo_2026}. Both routes must separate manipulation from camera and appearance changes and connect the learned signals to executable robot behavior.

\subsubsection{Latent Action Learning}\label{sec_taxonomy_latent_action}
Latent action learning encodes visual transitions or temporally extended behaviors into learned variables that serve as policy-training targets or control conditions. These variables have no predefined physical interpretation; geometric or semantic supervision may nevertheless constrain what they encode. We examine how the latent interface is learned, which changes it encodes, and how it is grounded in robot control.

\textbf{Learning latent action interfaces.}
A common approach combines an inverse model that encodes visual transitions with a forward model that predicts their outcomes. As a foundational approach, LAPO~\cite{schmidt2024learningactactions} learns latent action policies from action-free videos through inverse and forward dynamics modeling and grounds them in control using limited action labels. Genie~\cite{bruce2024geniegenerativeinteractiveenvironments} learns discrete latent actions and action-conditioned video dynamics from unlabeled videos to generate controllable interactive environments. Moto~\cite{chen2025motolatentmotiontoken} tokenizes video motion into latent codes, pretrains an autoregressive motion predictor, and jointly fine-tunes motion prediction and robot action generation. LAPA~\cite{ye_latent_2025} learns discrete transition codes through a VQ-VAE and uses them as prediction targets for VLA pretraining. Robot adaptation replaces the latent-action head with a robot-action head; the human-video experiments use Something-Something V2, providing evidence from general human manipulation videos. IGOR~\cite{chen_igor_2024} learns image-goal representations from mixed human and robot videos, including Ego4D and EPIC-KITCHENS, and uses the resulting codes to supervise policy and world-model learning. In contrast to using codes primarily as pretraining targets, UniSkill~\cite{kim_uniskill_2025} retains a latent skill interface during execution: skills extracted from human video prompts condition a policy trained on robot demonstrations. Its skill-learning data combine human videos, including H2O, with robot videos.

Some latent interfaces retain explicit motion structure. MimicPlay~\cite{wang_mimicplay_2023} learns latent plans through goal-conditioned prediction of 3D hand trajectories reconstructed from calibrated cameras, then trains a plan-conditioned robot controller. AMPLIFY~\cite{collins_amplify_2025} compresses tracked point velocities into motion tokens, predicts these tokens from observations and task descriptions, and maps them to robot actions through a separately trained inverse model. Both retain geometric or tracking supervision rather than learning solely from visual reconstruction.

\textbf{Making latent actions action-centric.}
Visual prediction alone can encode camera motion and appearance changes that are irrelevant to control. Bu et al.~\cite{bu2025univlalearningacttaskcentric} proposed UniVLA, which conditions latent action learning on language to suppress task-irrelevant visual change and support adaptation across robot embodiments. MVP-LAM~\cite{lee_mvp-lam_2026} addresses viewpoint dependence through cross-view reconstruction, i.e., a latent action inferred from one view must predict future visual features in another synchronized view. Training combines multi-view robot recordings with EgoExo4D, and the resulting tokens supervise VLA pretraining. Motion-Focused Latent Action~\cite{xu_motion-focused_2026} instead uses hand-segmentation masks to separate motion-related and background codes in EgoDex videos. Its motion codes supervise action-intent prediction, which is combined with current visual and proprioceptive information during robot adaptation.

ConLA~\cite{dai_conla_2026} combines action-category contrastive supervision with temporal reversal to shape latent actions, then uses them for policy pretraining; its human-video experiments use Something-Something V2. UniLACT~\cite{govind_unilact_2026} introduces depth-aware latent actions through joint RGB-depth modeling, although its reported training and evaluation focus on robot data. Across this category, the required supervision may include synchronized views, segmentation masks, action categories, or depth estimates.

\textbf{Grounding latent actions for robot control.}
Robot supervision can shape the latent interface itself, rather than being introduced only after video pretraining. GR00T N1~\cite{nvidia2025gr00tn1openfoundation} assigns latent action labels to egocentric human videos and incorporates them alongside real and synthetic robot data for generalist humanoid policy pretraining. villa-X~\cite{chen_villa-x_2025} augments visual prediction with robot state and action forecasting, and jointly models latent plans and robot actions so that the former condition the latter. CLAP~\cite{zhang_clap_2026} instead learns a token vocabulary from robot action sequences and contrastively aligns visual transitions with this vocabulary. Human videos, including Ego4D, thereby contribute pseudo-action targets for joint policy training. Robot trajectories support continuous-action post-training. HARP-VLA~\cite{zhu_harp-vla_2026} combines unpaired videos, including HOI4D~\cite{liu_hoi4d_2024} and OpenEgo~\cite{jawaid_openego_2025}, with paired human-robot demonstrations. Temporal alignment and manipulation-related auxiliary cues shape its visual and latent-action representations before robot-action fine-tuning. For these methods, transfer to new tasks and embodiments depends on the range of robot actions used for training and the reliability of the learned video-to-action mapping.

\textbf{Discussion.}
Latent actions can provide policy-training targets for videos without recorded robot commands, although their construction or grounding may still use robot-action supervision. The observed visual change can still correspond to several robot actions. Camera motion, occlusion, and unobserved contact dynamics can produce ambiguous codes. Similar visual outcomes may also require different actions across embodiments. Mapping these codes to robot control may still require action or alignment supervision. Evaluation should combine action-prediction probes and viewpoint-robustness tests with downstream task success and robot-data efficiency~\cite{lee_mvp-lam_2026, chen_villa-x_2025}. Mixed-data training also requires ablations that isolate the contribution of EHV.

\subsubsection{World-Action Modeling}\label{sec_taxonomy_world_action}

\begin{figure}[htbp]
\centering
\includegraphics[width=\textwidth]{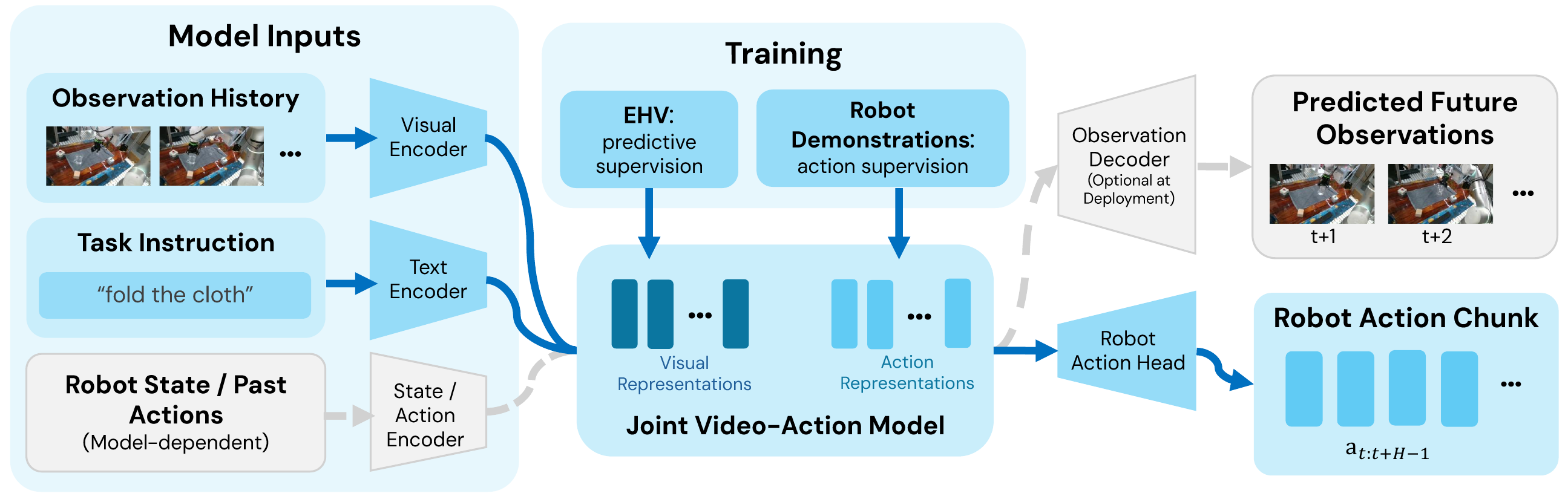}
\caption{A representative joint video-action architecture conditioned on visual observations and task instructions, with model-dependent robot state or past-action inputs. In the illustrated training setup, EHV supplies predictive supervision and robot demonstrations supply action supervision. Future-observation generation may be omitted at deployment. All video frames are from RH20T~\cite{fang_rh20t_2024}.}
\label{fig:world_action_modeling}
\end{figure}

World-action models predict future observations or interaction states for robot learning and control. Their uses include policy learning, action selection, and training-data generation. Figure~\ref{fig:world_action_modeling} illustrates a representative joint video-action architecture rather than a universal design. We organize the methods by these downstream roles, which can coexist within the same framework.

\textbf{Predictive pretraining and policy learning.}
Future prediction can provide a learning objective that connects human interaction videos to robot action generation. Wu et al.~\cite{wu_unleashing_2023} introduced GR-1, which pretrains on Ego4D video prediction and subsequently learns robot actions together with future images. GR-2~\cite{cheang_gr-2_2024} scales video pretraining on Internet videos, including egocentric sources, and then fine-tunes video generation and action prediction on robot trajectories. AffordanceWAM~\cite{you2026affordancewamaffordanceawarejointworldaction} augments joint video-action prediction with future affordance fields and heatmaps as shared cross-embodiment targets. Human videos supervise world prediction. Robot demonstrations additionally supervise action generation. Motion supervision links prediction to control: Ego-PM~\cite{zhang_ego-centric_2025} predicts hand trajectories and conditions future-frame generation on them, with evaluations in both human and robot settings. RynnVLA-001~\cite{jiang_rynnvla-001_2025} combines egocentric video pretraining with wrist-trajectory prediction before adapting to robot actions through a robot-specific ActionVAE. During deployment, it predicts action embeddings without generating future visual tokens. ViPRA~\cite{routray_vipra_2026} jointly predicts future observations and motion-centric latent actions from Something-Something~\cite{Goyal_2017_ICCV} and robot videos. Robot-supervised training then maps these latent actions to continuous actions through a decoder.

\looseness=-1
Policies can use predictive features without generating explicit future rollouts during execution. JoyAI-RA 0.5~\cite{team_joyai-ra_2026} pretrains a latent-action-conditioned world model on heterogeneous videos, including EHV, then uses its frozen features alongside visual-language features to condition an action expert. LaWAM~\cite{chen_lawam_2026} learns from mixed egocentric human and robot videos and conditions action generation on latent visual subgoals predicted by a latent-action-conditioned world model, avoiding future pixel reconstruction. World Tokens~\cite{tang_world_2026} instead shares conditioning tokens between video denoising and action prediction, removing the video branch at deployment; its robot-data experiments illustrate the mechanism rather than an isolated EHV contribution. HumanScale~\cite{ma_humanscale_2026} examines the pretraining-data question under a fixed world-action modeling protocol, reporting benefits from curated EHV with additional supervision from hand-derived pseudo-action labels. Together, these methods show that predictive learning and action supervision can be combined without requiring online video generation. WATCH~\cite{park2026learningtransferableworldactionmodels} pretrains on task-paired human demonstrations by using one demonstration's video and actions to predict another demonstration's future frames and actions. After robot post-training, it supports both demonstration-conditioned and context-free policies.

\textbf{Prediction-guided planning and evaluation.}
Predictive models can also assess the consequences of candidate actions before execution. V-JEPA 2~\cite{assran2025vjepa2selfsupervisedvideo} combines action-free predictive pretraining with action-conditioned adaptation on robot interactions to support goal-directed planning in latent space. DreamDojo~\cite{gao_dreamdojo_2026} pretrains a world model on EHV using continuous latent actions, then adapts its conditioning interface to target-robot actions. Its simulated rollouts support policy evaluation and action selection through an external value model. $\tau_0$-WM~\cite{zhou__0-wm_2026} combines joint video-action prediction with an action-conditioned simulator. EHV contributes video supervision, and robot data supplies executable action targets. At inference, re-denoising consistency ranks candidates, and low-quality candidates trigger simulator rollout and action refinement.

Moreover, other planning methods attempt to avoid dense video-based evaluation. CAPE~\cite{chen_cape_2026}, trained on DROID~\cite{khazatsky_droid_2025}, predicts action-conditioned future latent trajectories in parallel and compares them with goal representations for model predictive control. NovaFlow~\cite{li_novaflow_2025} instead converts generated task videos into 3D object flow, which guides trajectory optimization or deformable-object planning. These methods guide planning through predicted latent states or object motion, reducing reliance on evaluating full video rollouts.

\textbf{Predictive simulation and data generation.}
Controllable interaction prediction can generate examples for policy training. EgoHOI~\cite{li_egocentric_2026} conditions egocentric video synthesis on hand kinematics and head motion, using geometric priors to stabilize HOI without requiring future object trajectories. Its evaluation on HOT3D~\cite{Banerjee_2025_CVPR} assesses interaction-synthesis performance. Benefits for downstream robot learning remain to be evaluated. EgoGenesis~\cite{yan_egogenesis_2026-1} combines persistent 3D scene memory with geometry-aware motion conditioning and trains on mixed human, robot, and simulated data. Its generated trajectories augment real demonstrations in robot policy experiments. Generation must be controllable, and the resulting observations must remain aligned with their action labels.

\textbf{Discussion.}
A visually realistic prediction may still misrepresent action outcomes or depict behavior that the robot cannot execute. In EHV, camera motion, occlusion, and unobserved contact forces complicate dynamics learning. Embodiment differences also limit the direct transfer of human motion to robot control. Autoregressive rollouts can accumulate errors, but parallel prediction and candidate refinement do not eliminate model bias. Evaluation should therefore match the model's role, i.e., downstream task success and robot-data efficiency for predictive pretraining, action sensitivity and outcome-ranking accuracy for planning, and label consistency and policy improvement for synthetic data. Each comparison should also report inference cost and isolate the contribution of EHV through controlled ablations.

\subsubsection{Summary}
Latent action learning derives compact transition or skill representations that supervise policy pretraining or condition action generation~\cite{ye_latent_2025, kim_uniskill_2025}. World-action modeling learns predictive representations or future outcomes that support policy training, planning, evaluation, and synthetic-data generation~\cite{jiang_rynnvla-001_2025, gao_dreamdojo_2026, yan_egogenesis_2026-1}. The two routes can be combined by using latent actions to condition predictive models. Their deployment roles differ: predictive knowledge may be retained in policy features, accessed through online rollouts, or used only during training. These approaches can exploit videos without robot action labels, but may still require geometric priors, pseudo-action annotations, or robot demonstrations for grounding.

\textbf{Remark.} The learning paradigms leave different requirements for robot execution. For example, WARPED uses an additional workspace scan to construct robot-format training data~\cite{freeman_warped_2026}. R3M supplies visual features to policies trained on robot demonstrations~\cite{nair_r3m_2022}. DreamDojo adapts a human-video world model to target-robot actions~\cite{gao_dreamdojo_2026}. Comparing these methods therefore requires considering the geometric inputs, robot experience, and downstream validation alongside the amount of human video.


\section{Training Data Landscape}\label{sec_main_2}
The Ego2Robot learning paradigms reviewed in Section~\ref{sec_main_1} place different demands on data coverage, supervision, and embodiment alignment. We examine three data sources: egocentric human video, robot demonstrations, and synthetic or simulated data. EHV captures human behavior and interaction structure. Robot demonstrations provide embodiment-specific observation-action supervision. Synthetic and simulated resources can augment scenes and labels that are difficult to obtain from real-world recordings. We compare these resources by scale, viewpoints, modalities, and supervision, emphasizing their roles and limitations in Ego2Robot learning.


\subsection{Egocentric Human Videos}\label{sec_main_2_1}

EHVs constitute the principal source of human manipulation data considered in this survey. Head- or body-worn cameras record interactions from viewpoints that move with the demonstrator. Beyond visual appearance, they encode object affordances, HOIs, temporal information, and task structure. Depending on their modalities and annotation granularity, EHV datasets provide supervision ranging from visual and temporal representations to interaction geometry and task semantics. We summarize representative resources according to language annotations, scale, viewpoint, modalities, scenario, and selected annotation types, as shown in Table~\ref{tab:ehv_resources}.

\begin{table*}[!t]
\centering
\caption{
Comparison of representative human video resources containing egocentric views.
LA denotes task descriptions or narrations, rather than action-class labels alone
(\Yes{} present; \No{} absent; \LAHumanIcon{}~human-authored text; \LAModelIcon{}~model-generated text; \LAUnknownIcon{}~insufficiently specified provenance).
\SurveyView{Ego}/\SurveyView{Exo}: egocentric/exocentric views.
Scale follows source-specific counting conventions; multiview totals may sum
synchronized camera streams and should not be interpreted as egocentric duration.
Dashes `-' indicate entries not tabulated.
Modality icons denote \ModDepth{} depth, \ModMono{} monochrome video,
\ModAudio{} audio, \ModIMU{} inertial measurements (IMU), and \ModGaze{} gaze.
Listed modalities and annotations may cover only subsets of a resource.
Annotation icons, in left-to-right order, denote
\AnnHand{}~hand geometry, \AnnObject{}~object geometry, \AnnTask{}~task structure, \AnnContact{}~physical contact, and \AnnSemantic{}~interaction semantics. 
This column summarizes only these five types; a dash marks no selected type tabulated. Geometry-source superscripts identify the main annotation pipeline and exclude calibration and quality checks.
OpenEgo aggregates six public datasets and overlaps with other listed resources.
}
\label{tab:ehv_resources}
\begingroup
\newcommand{\EHVGeometrySlot}[2]{%
  \mbox{\fontsize{7.5}{8.5}\selectfont\makebox[1.9em][l]{%
    #1\kern-.15em\textsuperscript{\fontsize{5}{5}\selectfont#2}}}%
}
\newcommand{\EHVAnnotations}[5]{%
  \mbox{%
    \if\relax\detokenize{#1}\relax\else\EHVGeometrySlot{\AnnHand{}}{#1}\fi%
    \if\relax\detokenize{#2}\relax\else\EHVGeometrySlot{\AnnObject{}}{#2}\fi%
    \ifnum#3=1\relax\AnnotationSlot{\AnnTask{}}\fi%
    \ifnum#4=1\relax\AnnotationSlot{\AnnContact{}}\fi%
    \ifnum#5=1\relax\AnnotationSlot{\AnnSemantic{}}\fi%
  }%
}
\tablestyle{2.5pt}{1.05}
\fontsize{8}{9}\selectfont
\resizebox{0.98\linewidth}{!}{%
\setlength{\extrarowheight}{1.5pt}
\begin{tabular}{@{}M{28.5mm} C{10mm} C{9mm} C{11mm} C{15.5mm} C{16mm} M{42mm} C{24mm}@{}}
\toprule
\multirow{2}{*}{\textbf{Resource}} &
\multirow{2}{*}{\textbf{LA}} &
\multicolumn{2}{c}{\textbf{Scale}} &
\multirow{2}{*}{\textbf{Viewpoint}} &
\multirow{2}{*}{\shortstack{\textbf{Extra}\\\textbf{modalities}}} &
\multirow{2}{*}{\textbf{Scenario}} &
\multirow{2}{*}{\shortstack{\textbf{Annotation}\\\textbf{types}}} \\
\cmidrule(lr){3-4}
& & \textbf{Hours} & \textbf{Frames} & & & & \\
\midrule

\textbf{ARCTIC}~\cite{fan_arctic_2023} &
\LANone{} &
- & 2.1M &
\SurveyView{Ego + Exo} &
- &
Dexterous bimanual manipulation &
\EHVAnnotations{\textdagger}{\textdagger}{0}{1}{0} \\
\addlinespace[2pt]

\textbf{Assembly101}~\cite{sener_assembly101_2022} &
\LANone{} &
513 & 111M &
\SurveyView{Ego + Exo} &
\ModMono &
Toy-vehicle assembly and disassembly &
\EHVAnnotations{\textdaggerdbl}{}{1}{0}{1} \\
\addlinespace[2pt]

\textbf{CaptainCook4D}~\cite{peddi2024captaincook4ddatasetunderstandingerrors} &
\LAHuman{} &
94.5 & - &
\SurveyView{Ego} &
\ModDepth\ModAudio\ModIMU &
Cooking procedures with normal and erroneous execution &
\EHVAnnotations{*}{}{1}{0}{1} \\
\addlinespace[2pt]

\textbf{Ego4D}~\cite{grauman_ego4d_2022} &
\LAHuman{} &
3{,}700 & - &
\SurveyView{Ego} &
\ModAudio\ModIMU\ModGaze &
Daily activities &
\mbox{\EHVGeometrySlot{\AnnObject{}}{\S}\AnnotationSlot{\AnnSemantic{}}} \\
\addlinespace[2pt]

\textbf{EgoDex}~\cite{hoque_egodex_2026} &
\LAMixed{} &
829 & 90M &
\SurveyView{Ego} &
- &
Dexterous tabletop manipulation &
\EHVAnnotations{*}{}{0}{0}{0} \\
\addlinespace[2pt]

\textbf{EgoLive}~\cite{li_egolive_2026} &
\LAModel{} &
1{,}680 & - &
\SurveyView{Ego} &
\ModIMU &
Home service, retail, and pharmacy &
\mbox{\EHVGeometrySlot{\AnnHand{}}{\textdaggerdbl}\AnnotationSlot{\AnnTask{}}\AnnotationSlot{\AnnSemantic{}}} \\
\addlinespace[2pt]

\textbf{EgoVerse}~\cite{punamiya_egoverse_2026} &
\LAUnknown{} &
1{,}300 & - &
\SurveyView{Ego} &
- &
Manipulation in lab and real-world settings &
\EHVAnnotations{M}{}{0}{0}{0} \\
\addlinespace[2pt]

\textbf{EgoXtreme}~\cite{yoon_egoxtreme_2026} &
\LANone{} &
12.9 & 1.3M &
\SurveyView{Ego} &
- &
Maintenance, sports, and rescue (extreme visual conditions) &
\EHVAnnotations{}{\textdagger}{0}{0}{0} \\
\addlinespace[2pt]

\textbf{Ego-Exo4D}~\cite{grauman_ego-exo4d_2024} &
\LAHuman{} &
1{,}286 & - &
\SurveyView{Ego + Exo} &
\ModAudio\ModMono\ModIMU\ModGaze &
Skilled activities &
\mbox{\EHVGeometrySlot{\AnnHand{}}{\textdaggerdbl\S}\AnnotationSlot{\AnnTask{}}\AnnotationSlot{\AnnSemantic{}}} \\
\addlinespace[2pt]

\textbf{EgoExoLearn}~\cite{huang2025egoexolearndatasetbridgingasynchronous} &
\LAHuman{} &
120 & - &
\SurveyView{Ego + Exo} &
\ModGaze &
Demonstration following in daily life and laboratories &
\AnnTypes{0}{0}{1}{0}{1} \\
\addlinespace[2pt]

\mbox{\textbf{EPIC-KITCHENS}} \cite{damen_rescaling_2020} &
\LAHuman{} &
100 & 20M &
\SurveyView{Ego} &
\ModAudio &
Kitchen activities &
\AnnTypes{0}{0}{0}{0}{1} \\
\addlinespace[2pt]

\textbf{H2O}~\cite{kwon_h2o_2021} &
\LANone{} &
- & 571K &
\SurveyView{Ego + Exo} &
\ModDepth &
Bimanual object manipulation &
\EHVAnnotations{\textdaggerdbl}{\textdaggerdbl}{0}{0}{0} \\
\addlinespace[2pt]

\textbf{HD-EPIC}~\cite{perrett_hd-epic_2025} &
\LAHuman{} &
41.3 & 4.46M &
\SurveyView{Ego} &
\ModAudio\ModGaze &
Unscripted cooking &
\EHVAnnotations{}{\textdaggerdbl\S}{1}{0}{1} \\
\addlinespace[2pt]

\textbf{HO-Cap}~\cite{wang2025hocapcapturedataset3d} &
\LANone{} &
- & 656K &
\SurveyView{Ego + Exo} &
\ModDepth &
Object use, pick-and-place, and handovers &
\EHVAnnotations{\textdaggerdbl}{\textdaggerdbl}{0}{0}{0} \\
\addlinespace[2pt]

\textbf{HOI4D}~\cite{liu_hoi4d_2024} &
\LANone{} &
- & 2.4M &
\SurveyView{Ego} &
\ModDepth &
Object interaction across categories &
\EHVAnnotations{\textdaggerdbl\S}{\textdaggerdbl\S}{0}{0}{0} \\
\addlinespace[2pt]

\textbf{HoloAssist}~\cite{wang_holoassist_2023} &
\LAHuman{} &
166 & 17.9M &
\SurveyView{Ego} &
\ModDepth\ModAudio\ModIMU\ModGaze &
Guided manipulation and procedural tasks &
\EHVAnnotations{*}{}{1}{0}{1} \\
\addlinespace[2pt]

\textbf{HOT3D}~\cite{Banerjee_2025_CVPR} &
\LANone{} &
13.9 & 3.7M &
\SurveyView{Ego} &
\ModMono\ModGaze &
Household and office object manipulation &
\EHVAnnotations{\textdagger}{\textdagger}{0}{0}{0} \\
\addlinespace[2pt]

\textbf{IndustReal}~\cite{schoonbeek2023industrealdatasetprocedurestep} &
\LANone{} &
5.8 & - &
\SurveyView{Ego} &
\ModDepth\ModMono\ModGaze &
Industrial-like assembly with procedural and execution errors &
\EHVAnnotations{*}{\S}{1}{0}{1} \\
\addlinespace[2pt]

\textbf{Nymeria}~\cite{ma_nymeria_2024} &
\LAHuman{} &
300 & 201.2M &
\SurveyView{Ego + Exo} &
\ModAudio\ModMono\ModIMU\ModGaze &
Unscripted daily activities &
\AnnTypes{0}{0}{0}{0}{0} \\
\addlinespace[2pt]

\textbf{OakInk2}~\cite{zhan_oakink2_2024} &
\LAMixed{} &
- & 4M &
\SurveyView{Ego + Exo} &
- &
Bimanual manipulation with complex goals &
\EHVAnnotations{\textdagger}{\textdagger}{1}{0}{1} \\
\addlinespace[2pt]

\textbf{OpenEgo}~\cite{jawaid_openego_2025} &
\LAModel{} &
1{,}100 & 119.6M &
\SurveyView{Ego} &
- &
Source-dependent activities &
\EHVAnnotations{M}{}{0}{0}{1} \\
\addlinespace[2pt]

\textbf{Open-AoE}~\cite{li_open-aoe_2026} &
\LAMixed{} &
2{,}000 & - &
\SurveyView{Ego} &
- &
Everyday manipulation in natural settings &
\EHVAnnotations{\textdaggerdbl}{}{0}{0}{1} \\
\bottomrule
\end{tabular}
}

\par\smallskip
\noindent\parbox{0.98\linewidth}{\raggedright
Geometry sources:
\textsuperscript{*}~device-provided tracking;
\textsuperscript{\textdagger}~marker-based motion capture;
\textsuperscript{\textdaggerdbl}~reconstruction/model estimation;
\textsuperscript{\S}~manual geometry/CAD;
\textsuperscript{M}~mixed/inherited sources.
Multiple marks indicate combined sources.}
\endgroup

\end{table*}

The resources in Table~\ref{tab:ehv_resources} differ primarily in the type of supervision. EPIC-KITCHENS~\cite{damen_rescaling_2020}, Ego4D~\cite{grauman_ego4d_2022}, and Assembly101~\cite{sener_assembly101_2022} offer large-scale visual and temporal observations that support representation learning, interaction understanding, and task-progression modeling. Resources such as H2O~\cite{kwon_h2o_2021}, HOI4D~\cite{liu_hoi4d_2024}, ARCTIC~\cite{fan_arctic_2023}, and EgoDex~\cite{hoque_egodex_2026} provide more explicit geometric information about hands, objects, motion, or contact, which is useful for correspondence estimation and cross-embodiment alignment. Ego-Exo4D~\cite{grauman_ego-exo4d_2024} additionally introduces multi-view observations. Datasets with language annotations provide semantic supervision that links observed behavior to task-level verbs, nouns, and instructions. The utility of an EHV resource depends not only on its scale, but also on how well its available supervision matches the intended Ego2Robot objective.

Their varied objects, environments, and interactions provide visual evidence for learning perceptual, geometric, temporal, and semantic priors. However, most EHVs do not contain executable robot action labels, and differences in embodiment, kinematics, sensing configuration, and viewpoint create gaps between human demonstrations and robot observations. Errors introduced by pose estimation or reconstruction may further propagate to downstream learning. Using these priors for control still requires a mapping from human observations to the robot's representations and action space.

\subsection{Robot Demonstration Data}\label{sec_main_2_2}

Robot demonstration data pair observations with embodiment-specific states and actions, supporting policy learning, action-space calibration, and downstream validation. This section also includes paired human-robot resources and handheld manipulation interfaces that record robot-compatible motion rather than complete robot executions. YUBI~\cite{ohkawa_yubi_2026} collects bimanual demonstrations with tracked handheld grippers; these devices are the collection interface. Robot platforms are used for downstream evaluation. Table~\ref{tab:robot_demo_resources} compares these resources by embodiment or interface, scale, observation-action information, contact sensing, and human-robot associations.

\begin{table*}[!t]
\centering

\caption{
Comparison of representative robot demonstration resources.
Multi: multiple robot embodiments; ep.: episodes; traj.: trajectories;
env.: environments; h: hours.
Obs./Act.: observations/actions; EE: end effector; proprio.: proprioception.
Contact sensing: F/T, force/torque; T, tactile; T\textsuperscript{*}, tactile on a subset;
Mixed, heterogeneous coverage; NR, not reported.
H--R pairs: \Yes{} human-robot demonstration associations; \No{} no pairing. Associations need not be frame-aligned or one-to-one. YUBI uses handheld grippers.
}
\label{tab:robot_demo_resources}

\tablestyle{2.5pt}{1.03}
\fontsize{8}{9}\selectfont
\resizebox{0.98\linewidth}{!}{%
\begin{tabular}{@{}M{33mm} M{20mm} M{28mm} M{48mm} C{15mm} C{11mm}@{}}
\toprule
\textbf{Resource} &
\textbf{Embodiment}\newline\textbf{/interface} &
\textbf{Scale} &
\textbf{Observations and actions} &
\textbf{Contact}\newline\textbf{sensing} &
\textbf{H--R}\newline\textbf{pairs} \\
\midrule

\textbf{RT-1}~\cite{brohan_rt-1_2023} &
Single-arm &
130K ep.; $>$700 tasks &
\textbf{Obs:} RGB, language\newline
\textbf{Act:} arm, base, gripper &
NR &
\No{} \\

\textbf{BridgeData V2}~\cite{walke_bridgedata_nodate} &
Single-arm &
60K traj.; 24 env. &
\textbf{Obs:} Multi-view RGB/RGB-D\newline
\textbf{Act:} Cartesian EE, gripper &
NR &
\No{} \\

\textbf{DROID}~\cite{khazatsky_droid_2025} &
Single-arm &
76K successful + 16K failed traj. &
\textbf{Obs:} Multi-view RGB-D, proprio.\newline
\textbf{Act:} joint/EE, gripper &
NR &
\No{} \\

\textbf{AIRoA MoMa}~\cite{takanami_airoa_2025} &
Single-arm &
25K ep.; $\sim$94 h &
\textbf{Obs:} Dual-view RGB, robot states\newline
\textbf{Act:} arm, base, gripper &
F/T &
\No{} \\
\midrule

\textbf{Open X-Embodiment}~\cite{collaboration_open_2025} &
Multi &
1M+ traj.; 160K tasks; 527 skills &
\textbf{Obs:} Heterogeneous vision/proprio.\newline
\textbf{Act:} heterogeneous robot actions &
Mixed &
\No{} \\

\textbf{RoboCOIN}~\cite{wu_robocoin_2026} &
Multi &
180K ep.; 421 tasks &
\textbf{Obs:} Multi-view RGB, kinematics\newline
\textbf{Act:} bimanual &
NR &
\No{} \\

\textbf{RoboMIND 2.0}~\cite{hou_robomind_2026} &
Multi &
310K traj.; 739 tasks &
\textbf{Obs:} Multi-view RGB-D, proprio.\newline
\textbf{Act:} bimanual/mobile &
T\textsuperscript{*} &
\No{} \\

\textbf{VTouch++}~\cite{hua_vtouch_2026} &
Multi &
120K+ ep.; 380+ tasks; $>$1{,}000 h &
\textbf{Obs:} Multi-view RGB-D, proprio.\newline
\textbf{Act:} bimanual &
T &
\No{} \\

\textbf{RH20T}~\cite{fang_rh20t_2024} &
Multi &
110K+ seq. &
\textbf{Obs:} RGB-D, audio, proprio.\newline
\textbf{Act:} robot actions &
F/T + T\textsuperscript{*} &
\Yes{} \\
\midrule

\textbf{BRMData}~\cite{zhang_empowering_2024} &
Bimanual &
500 traj.; 10 tasks &
\textbf{Obs:} Multi-view RGB-D, robot states\newline
\textbf{Act:} bimanual/base &
F/T &
\No{} \\

\textbf{Galaxea Open-World}~\cite{jiang_galaxea_2025} &
Bimanual &
100K traj.; 150 task categories; $>$500 h &
\textbf{Obs:} Head RGB, wrist RGB-D, proprio.\newline
\textbf{Act:} dual-arm/base &
NR &
\No{} \\

\textbf{YUBI}~\cite{ohkawa_yubi_2026} &
Bimanual &
1.2M ep.; 119 tasks; 8.4K h &
\textbf{Obs:} RGB, 6-DoF gripper tracking\newline
\textbf{Act:} pose/jaw &
NR &
\No{} \\

\textbf{ABC130K}~\cite{allshire_scalable_2026} &
Bimanual &
130K ep.; 195 tasks; 3.5K h; 400+ h (simulated) &
\textbf{Obs:} Three-view RGB, robot states\newline
\textbf{Act:} bimanual joint/gripper &
NR &
\No{} \\
\midrule

\textbf{AgiBot World Colosseo}~\cite{agibot-world-contributors_agibot_2025} &
Humanoid &
1M traj.; 217 tasks; 106 scenarios &
\textbf{Obs:} Multi-view RGB-D, proprio.\newline
\textbf{Act:} whole-body/bimanual &
T\textsuperscript{*} &
\No{} \\

\textbf{Humanoid Everyday}~\cite{zhao_humanoid_2026} &
Humanoid &
10.3K traj.; 260 tasks; $>$3M frames &
\textbf{Obs:} RGB-D, LiDAR, IMU, joint states\newline
\textbf{Act:} full-body joints &
T &
\No{} \\

\textbf{RoboTacDex}~\cite{wang_robotacdex_nodate} &
Humanoid &
6K traj.; 19 tasks &
\textbf{Obs:} Multi-view RGB-D, robot states\newline
\textbf{Act:} bimanual/dexterous &
T &
\No{} \\
\midrule

\textbf{HRDexDB}~\cite{lim_hrdexdb_2026} &
Multi &
3.2K trials (541 human and 2{,}674 robot trials in v3) &
\textbf{Obs:} Multi-view/ego RGB, kinematics\newline
\textbf{Act:} grasp trajectories &
T\textsuperscript{*} &
\Yes{} \\
\bottomrule
\end{tabular}
}

\end{table*}

Real-robot datasets differ primarily in their scale and coverage of robot embodiments. RT-1~\cite{brohan_rt-1_2023} demonstrates the feasibility of large-scale data collection across a fleet of real robots. Open X-Embodiment~\cite{collaboration_open_2025} and RoboMIND 2.0~\cite{hou_robomind_2026} place greater emphasis on learning across different robot configurations and morphologies. DROID~\cite{khazatsky_droid_2025} and AgiBot World Colosseo~\cite{agibot-world-contributors_agibot_2025}, meanwhile, broaden the coverage of real-world environments, tasks, and interaction scenarios. Some datasets also record force or tactile measurements. AIRoA MoMa~\cite{takanami_airoa_2025} records six-axis wrist force--torque measurements, RoboTacDex~\cite{wang_robotacdex_nodate} provides visual and tactile feedback, and RoboMIND 2.0~\cite{hou_robomind_2026} includes a dedicated tactile-augmented subset. Such signals are particularly valuable for contact-rich manipulation, precision insertion, and dexterous grasping, where relevant interaction states may be difficult to infer from vision alone. Nevertheless, force and tactile measurements are not yet standard components of large-scale robot demonstration datasets.

Robot demonstration data provide executable action supervision, measured internal states, and task outcomes that can ground information learned from EHV. Their collection remains sensitive to platform, task, and environment. Differences in counting conventions and dataset overlap preclude a universal comparison of scale or diversity with human video. Within Ego2Robot learning, these resources support action alignment, policy adaptation, and real-world validation alongside EHV.

\subsection{\texorpdfstring{\protect\raisebox{2ex}[0pt][0pt]{\protect\hypertarget{milestone.synthetic-data}{}}}{}Synthetic and Simulated Data}\label{sec_main_2_3}

Synthetic and simulated data are generated through physics-based simulation, demonstration transformation, real-to-sim reconstruction, generative video models~\cite{chen_hvg3d_2026}, or embodiment-aware editing. The resources reviewed here augment EHV or real-robot demonstrations, support cross-domain alignment, or provide downstream validation. Their value lies in expanding scene and trajectory coverage and, where available, providing simulator states, action labels, or structural annotations. Table~\ref{tab:synthetic_resources} summarizes representative resources in terms of their generation methods, required inputs, generated content, and available supervision.
\begin{table*}[!t]
\centering

\caption{
Comparison of representative synthetic and simulated data resources, grouped by generation method or resource type.
Demos: demonstrations; EHV: egocentric human video; sim.: simulation.
The Validation columns distinguish resource use from the evaluation setting.
Under \emph{Use}, Train denotes policy training, Plan denotes model-based planning,
and Benchmark denotes policy evaluation.
Under \emph{Eval.}, Sim and Real denote evaluation in simulation and on
physical robots, respectively.
NR indicates information not reported.
}
\label{tab:synthetic_resources}

\tablestyle{2.3pt}{1.03}
\fontsize{8}{9}\selectfont
\resizebox{0.98\linewidth}{!}{%
\begin{tabular}{@{}M{32mm} M{32mm} M{30mm} M{33mm} C{15mm} C{13mm}@{}}
\toprule
\multirow{2}{*}{\textbf{Resource}} &
\multirow{2}{*}{\textbf{Input}} &
\multirow{2}{*}{\textbf{Output}} &
\multirow{2}{*}{\textbf{Supervision}} &
\multicolumn{2}{c}{\textbf{Validation}} \\
\cmidrule(lr){5-6}
& & & & \textbf{Use} & \textbf{Eval.} \\
\midrule
\multicolumn{6}{@{}l}{\textit{Simulation synthesis}} \\
\addlinespace[1pt]

\textbf{CALVIN}~\cite{mees_calvin_2022} &
Robot demos &
Tasks + trajectories &
State + action &
Train & Sim \\

\textbf{GenSim}~\cite{wang_gensim_2024} &
Task specification &
Tasks + trajectories &
Simulator labels &
Train & Sim + Real \\

\textbf{RoboGen}~\cite{wang_robogen_nodate} &
Task specification &
Tasks + trajectories &
Simulator labels &
Train & Sim \\

\textbf{RoboCerebra}~\cite{han_robocerebra_2025} &
Task specification &
Tasks + trajectories &
State + action &
Train & Sim \\

\textbf{Nimbus}~\cite{he_nimbus_2026} &
Task specification &
Tasks + trajectories &
Simulator labels &
NR & NR \\
\midrule
\multicolumn{6}{@{}l}{\textit{Demo augmentation}} \\
\addlinespace[1pt]

\textbf{MimicGen}~\cite{mandlekar_mimicgen_2023} &
Robot demos &
Augmented trajectories &
Recomputed actions after trajectory transformation &
Train & Sim + Real \\

\textbf{SoftMimicGen}~\cite{moghani_softmimicgen_2026} &
Robot demos &
Augmented trajectories &
Transformed actions &
Train & Sim + Real \\

\textbf{SkillMimicGen}~\cite{garrett_skillmimicgen_2024} &
Robot demos &
Augmented trajectories &
Adapted skill actions + planned motions &
Train & Sim + Real \\
\midrule
\multicolumn{6}{@{}l}{\textit{Real-to-sim}} \\
\addlinespace[1pt]

\textbf{RoboTwin}~\cite{mu_robotwin_2025} &
Object images + spatial annotations + task spec. &
Sim. scenes + trajectories &
State + action &
Train & Sim + Real \\

\textbf{RoboSnap}~\cite{zhang_robosnap_2026} &
Real observation &
Sim. scenes &
Simulator labels &
Train & Sim + Real \\

\textbf{WANDA}~\cite{guo_worlds_2026} &
Robot demos &
Augmented trajectories &
State + action &
Train & Sim + Real \\
\midrule
\multicolumn{6}{@{}l}{\textit{Video generation}} \\
\addlinespace[1pt]

\textbf{TASTE-Rob pipeline}~\cite{zhao_taste-rob_2025} &
Initial image + instruction &
Generated HOI videos &
Visual supervision &
Train & Sim \\

\textbf{EgoSim}~\cite{hao_egosim_2026} &
Human videos + trajectories &
Interaction videos &
State + action &
Plan & Real \\

\textbf{EgoDemoGen}~\cite{xu_egodemogen_2026} &
Robot demos &
Novel-view videos + transferred trajectories &
View-consistent observations and transformed actions &
Train & Sim + Real \\
\midrule
\multicolumn{6}{@{}l}{\textit{Evaluation benchmark}} \\
\addlinespace[1pt]

\textbf{REALM}~\cite{sedlacek_realm_2025} &
Real scenes + simulated counterparts &
Perturbed evaluation environments &
Task definitions and evaluation conditions &
Benchmark & Sim + Real \\
\midrule
\multicolumn{6}{@{}l}{\textit{Embodiment editing}} \\
\addlinespace[1pt]

\textbf{HandEdit}~\cite{yang_handedit_2026} &
EHV + robot model &
Edited observations &
Paired editing &
NR & NR \\
\bottomrule
\end{tabular}
}

\end{table*}

Existing approaches can be grouped into three categories according to their data-generation pipelines. The first includes resources in which manipulation data are produced within simulation. CALVIN~\cite{mees_calvin_2022} provides teleoperated play trajectories for language-conditioned policy learning and long-horizon evaluation. GenSim~\cite{wang_gensim_2024} and RoboGen~\cite{wang_robogen_nodate} use large language models to automate task construction and support data generation. RoboCerebra~\cite{han_robocerebra_2025} combines LLM-generated tasks and subtask decompositions with human-executed simulation trajectories and temporal annotations. LLMs automate task specification, but demonstration collection and verification can still require substantial human effort. The second uses limited real-world inputs to generate additional trajectories or simulation environments. This includes MimicGen~\cite{mandlekar_mimicgen_2023}, SkillMimicGen~\cite{garrett_skillmimicgen_2024}, SoftMimicGen~\cite{moghani_softmimicgen_2026}, RoboTwin~\cite{mu_robotwin_2025}, RoboSnap~\cite{zhang_robosnap_2026}, and WANDA~\cite{guo_worlds_2026}. These approaches employ demonstration-segment transfer, skill stitching, deformable-object simulation, digital-twin construction, or real-to-sim reconstruction to reconstruct or diversify scenes, object configurations, and trajectories. The third directly generates or edits visual observations, as exemplified by TASTE-Rob~\cite{zhao_taste-rob_2025}, EgoDemoGen~\cite{xu_egodemogen_2026}, EgoSim~\cite{hao_egosim_2026}, and HandEdit~\cite{yang_handedit_2026}.

Although simulation typically provides precise state and action information within the simulated environment, its utility remains constrained by the fidelity of contact models, the quality of simulation assets, and the embodiment gap. Synthetic and simulated resources are therefore particularly useful for supplementing supervision unavailable in EHV, extending the coverage of real demonstrations, and constructing controlled evaluations of generalization. Real-robot execution is still needed to test whether the resulting behaviors are physically feasible.

\section{Open Challenges}\label{sec_challenges}
Sections~\ref{sec_main_1} and~\ref{sec_main_2} show which information EHV provides and which requirements remain for robot execution. In this section, we discuss five future directions: agent coordination, in-context adaptation, physical feasibility, controlled evaluation, and predictive control. Each direction depends on what EHV omits and what robot data or execution feedback can supply.

\subsection{AI Agents for Ego2Robot Learning}

An AI agent could use EHV-derived task information to select robot skills and monitor their outcomes. VoLo~\cite{chen2026volophysicalorchestratoropenvocabulary} introduces physical orchestration, treating VLA/WAM policies, perception models, and action primitives as interruptible tools for planning, monitoring, and recovery. RoboClaw~\cite{li2026roboclawagenticframeworkscalable} unifies data collection, policy learning, and task execution, using Entangled Action Pairs to couple manipulation and recovery behaviors for autonomous data collection and policy refinement. These frameworks provide methodological context for Ego2Robot integration. EHV could supply subtask dependencies, expected object-state changes, and interaction constraints. The agent would then map this information to available robot skills and check the outcomes using execution feedback.

For each video-derived step, the agent needs to check whether its prerequisites hold in the current scene and whether the available skills can achieve a verifiable outcome. Memory and tool feedback should track achieved states and failed attempts, supporting timely interruption, recovery, or targeted data collection. With controllers and tool access fixed, evaluation can separate errors in EHV-derived planning and monitoring from failures of the underlying skills. Additional robot data and policy updates must be counted when the skill library itself improves.

\subsection{In-Context Adaptation from Demonstrations and Feedback}\label{sec_taxonomy_in_context}

In-context adaptation uses demonstrations or interaction history to change robot behavior without updating model parameters during deployment. We divided it into three context interfaces: demonstration conditioning, as in GPT-Policy and WATCH~\cite{cheng2026incontextrobotlearningvlm, park2026learningtransferableworldactionmodels}; execution-history and action-effect conditioning, as in Qwen-RobotManip and Zeva-Ego~\cite{yuan2026qwenrobotmaniptechnicalreportalignment, huang2026zevaegoegocentricmidtrainingincontext}; and retrieved context, as in Robota-Embed~\cite{liu2026robotaembedunifiedvlaembeddings}. These mechanisms require different context representations and training procedures. Fixed deployment weights do not imply that the complete system requires no training or robot experience.

Context must retain task-relevant action order and object-state changes and remain applicable to the current scene, embodiment, and control interface. EHV-derived task information must be distinguished from knowledge acquired through robot feedback. Matched, mismatched, and absent context should be tested with the same policy checkpoint. Reports should specify how the context interface was trained and give the context length, retrieval settings, and memory reset rules. For repeated-trial adaptation, performance should be reported against robot interaction budgets.

\subsection{Cross-Embodiment Alignment and Physical Feasibility}

The robot may need a different motion to preserve the task-relevant interaction. Occlusion, camera motion, and reconstruction errors can corrupt geometric supervision. Differences in morphology and actuation may make demonstrated configurations infeasible for the target robot. Related work on contact-consistent reconstruction and physics-informed retargeting addresses parts of this problem~\cite{ren_c2dex_2026, pan_spider_2026}, but its reliability remains dependent on observation quality and modeling assumptions. Learned features and intermediate targets also need to remain applicable to the target robot, even when the method does not retarget an explicit trajectory.

Alignment objectives should preserve functional outcomes and allow robot-specific solutions. Retargeting and policy learning should retain reconstruction uncertainty and use it when weighting human-derived supervision. Recording failures separately at reconstruction, retargeting, and execution would make these sources of error easier to distinguish.

Physical grounding also requires information that ordinary EHV may not determine, including contact forces, friction, and occluded contact states. Limited synchronized tactile or force-torque data can supply complementary supervision. Execution-time sensing can also correct video-derived priors. RoboPaint~\cite{fan_robopaint_2026} retargets human demonstrations using camera, glove, and tactile recordings, which require more instrumentation than monocular video. Methods should remain usable when a modality is missing and report uncertainty in the inferred physical state. Evaluation should include contact-rich task success, slip and excessive-force events, and the cost of additional sensing.

\subsection{Evaluation, Attribution, and Interpretability}

Robot experiments have shown benefits from EHV pretraining~\cite{nair_r3m_2022}, but evaluation should establish when these benefits occur and what information produces them. It should measure whether EHV reduces the robot demonstrations needed to reach a given success rate and whether gains persist on unseen objects and environments. Comparisons with equally sized non-egocentric video datasets can help determine whether first-person interaction content contributes beyond additional visual pretraining. Policy architecture, auxiliary supervision, and training budgets should remain fixed in these comparisons.

\looseness=-1
Diagnostic tests can examine object-state recognition, contact estimation, task progress, or procedural reasoning~\cite{li_egoprocevqa_2026}. Controlled changes to observations, instructions, or retrieved context can then test whether the policy uses this information. Robot trials should check whether the identified errors explain execution failures, such as a failed grasp caused by an incorrect contact estimate.

\subsection{Reliable Prediction and Closed-Loop Control}

Latent actions and world-action models connect visual transitions to behavior, but predictive accuracy does not necessarily imply controllability. Latent variables may encode camera motion or appearance changes rather than manipulation. Visually plausible rollouts may also misrepresent contacts or responses to robot commands. DreamDojo~\cite{gao_dreamdojo_2026} combines EHV pretraining with post-training on target-robot data to support action-conditioned simulation and model-based planning. Nevertheless, reliable control requires predictions to remain useful under action interventions, unfamiliar interactions, and deviations from the demonstrated trajectory.

Tests should change candidate actions and check whether the predicted outcomes change accordingly. They should also assess whether prediction uncertainty helps trigger replanning or requests for further observations. Closed-loop evaluation should include disturbances, recovery from failed contacts, and long-horizon error accumulation. For online rollouts, latency limits how quickly predictions can inform control decisions, and models used only during training should instead be assessed through the resulting policy.

\section{Conclusion}\label{sec_conclusion}
This paper presents Ego2Robot learning in four learning paradigms and reviews their supporting data resources. Video scale alone does not establish EHV's value, and the retained interaction information must support control of the target robot. Training-time transfer and deployment-time context offer complementary routes, i.e., human experience from EHV can shape representations, supervision, and predictive models, and guide execution through demonstrations and retrieved knowledge. Both uses of EHV depend on mapping human-derived information to feasible robot behavior and validating it through execution. For experiments, controlled comparisons should measure task success, robustness, and robot-data efficiency and account for robot experience, context, sensing, and computational cost.

\section{Acknowledgements}\label{sec_ack}
The research work described in this paper was conducted in the JC STEM Lab of Machine Learning and Computer Vision funded by The Hong Kong Jockey Club Charities Trust. This research received partial support from the Global STEM Professorship Scheme from the Hong Kong Special Administrative Region. This publication was partially supported by the Hong Kong Polytechnic University through the Undergraduate Research and Innovation Scheme.

\FloatBarrier
\bibliographystyle{assets/plainnat}
\bibliography{bibs/20260820-Survey_D,bibs/20260827-Motivation,bibs/20260909-RelatedSurvey,bibs/20260918-1746-PublishedArticles,bibs/20260919-2000-DiagramCite}

\end{document}